%% file: main.tex
\documentclass[10pt]{article} 
\usepackage[accepted]{tmlr}

\input{math_commands.tex}

\usepackage{hyperref}
\usepackage{url}

\newcommand{\ie}{{\em i.e.,}}
\newcommand{\eg}{{\em e.g.,}}
\newcommand{\Ni}{({\em i})~}
\newcommand{\Nii}{({\em ii})~}
\newcommand{\Niii}{({\em iii})~}

\usepackage{booktabs}
\usepackage{enumitem}
\usepackage{graphicx}
\usepackage{subfigure}
\usepackage{pgfplots}
\usepackage{tikz}
\usepackage{wrapfig}
\usepackage{fdsymbol}
\usepackage{listings}
\usepackage{caption}
\usepackage{float}
\usepackage{adjustbox}
\usepackage{tikz}
\usepackage{pgfplots}
\pgfplotsset{compat=1.18}
\usepackage[edges]{forest}
\usepackage{listings}
\usepackage{cleveref}
\usepackage{multirow}
\usepackage{arydshln}
\usepackage{makecell}
\usepackage{tikz}
\usepackage{collcell}
\usepackage{diagbox}
\usepackage{rotating}
\usepackage{pgf}
\usepackage{transparent}
\usepackage{pifont}
\newcommand{\cmark}{\ding{51}} 
\newtheorem{definition}{Definition}

\definecolor{framework-blue}{RGB}{47, 85, 151}
\definecolor{content-yellow}{RGB}{255, 230, 153}
\definecolor{framework-yellow}{RGB}{255, 255, 255}
\definecolor{content-orange}{RGB}{251, 229, 215}
\definecolor{framework-orange}{RGB}{248, 203, 175}
\definecolor{content-gray}{RGB}{237, 237, 237}
\definecolor{framework-gray}{RGB}{166, 166, 166}
\definecolor{paired-light-blue}{RGB}{198, 219, 239}
\definecolor{paired-dark-blue}{RGB}{49, 130, 188}
\definecolor{paired-light-orange}{RGB}{251, 208, 162}
\definecolor{paired-dark-orange}{RGB}{230, 85, 12}
\definecolor{paired-light-green}{RGB}{199, 233, 193}
\definecolor{paired-dark-green}{RGB}{49, 163, 83}
\definecolor{paired-light-purple}{RGB}{218, 218, 235}
\definecolor{paired-dark-purple}{RGB}{117, 107, 176}
\definecolor{paired-light-gray}{RGB}{217, 217, 217}
\definecolor{paired-dark-gray}{RGB}{99, 99, 99}
\definecolor{paired-light-pink}{RGB}{222, 158, 214}
\definecolor{paired-dark-pink}{RGB}{123, 65, 115}
\definecolor{paired-light-red}{RGB}{231, 150, 156}
\definecolor{paired-dark-red}{RGB}{131, 60, 56}
\definecolor{paired-light-yellow}{RGB}{231, 204, 149}
\definecolor{paired-dark-yellow}{RGB}{141, 109, 49}
\tikzset{%
    parent/.style = {
        align=center,
        text width=2.5cm,
        rounded corners=3pt,
        line width=0.5mm,
        draw=gray!80
    },
    domain_box/.style = {
        align=center,
        text width=2.5cm,
        rounded corners=3pt,
        draw=framework-gray,
        line width=0.5mm
    },
    models/.style = {
        align=justify,
        text width=5.5cm, 
        rounded corners=3pt,
        fill=paired-light-blue!45,
        draw=framework-gray,
        line width=0.3mm,
        font=\tiny, 
        inner sep=4pt
    }
}

\title{A Comprehensive Survey of Contamination Detection Methods in Large Language Models}

\author{\name Mathieu Ravaut \email mathieuj001@e.ntu.edu.sg \\
        \addr Nanyang Technological University\\
        Institute of Infocomm Research (I2R), A*STAR
        \AND
        \name Bosheng Ding \email bosheng001@e.ntu.edu.sg\\
        \addr Nanyang Technological University
        \AND
        \name Fangkai Jiao \email fangkai002@e.ntu.edu.sg\\
        \addr Nanyang Technological University\\
        Institute of Infocomm Research (I2R), A*STAR
        \AND
        \name Hailin Chen \email hailin001@e.ntu.edu.sg\\
        \addr Nanyang Technological University
        \AND 
        \name Xingxuan Li \email xingxuan001@e.ntu.edu.sg \\
        \addr Nanyang Technological University
        \AND 
        \name Ruochen Zhao \email ruochen002@e.ntu.edu.sg \\
        \addr Nanyang Technological University
        \AND 
        \name Chengwei Qin \email chengwei003@e.ntu.edu.sg \\
        \addr Nanyang Technological University
        \AND 
        \name Caiming Xiong \email cxiong@salesforce.com \\
        \addr Salesforce Research
        \AND 
        \name Shafiq Joty \email sjoty@salesforce.com \\
        \addr Nanyang Technological University\\
        Salesforce Research
}

\def\month{07}  
\def\year{2025} 
\def\openreview{\url{https://openreview.net/forum?id=SxNMjbtdFm}} 

\begin{document}

\maketitle

\begin{abstract}
With the rise of Large Language Models (LLMs) in recent years, abundant new opportunities are emerging, but also new challenges, among which contamination is quickly becoming critical. 
Business applications and fundraising in Artificial Intelligence (AI) have reached a scale at which a few percentage points gained on popular question-answering benchmarks could translate into dozens of millions of dollars, placing high pressure on model integrity.
At the same time, it is becoming harder and harder to keep track of the data that LLMs have seen; if not impossible with closed-source models like GPT-4 and Claude-3 not divulging any information on the training set. 
As a result, contamination becomes a major issue: LLMs' performance may not be reliable anymore, as the high performance may be at least partly due to their previous exposure to the data. 
This limitation jeopardizes real capability improvement in the field of NLP, yet, there remains a lack of methods on how to efficiently detect contamination.
In this paper, we survey all recent work on contamination detection with LLMs, analyzing their methodologies and use cases to shed light on the appropriate usage of contamination detection methods. 
Our work calls the NLP research community's attention into systematically taking into account contamination bias in LLM evaluation. 
\end{abstract}

\input{Sections/1_introduction}

\input{Sections/2_definitions}
\input{Sections/3_open_data_contamination}
\input{Sections/4_closed_data_contamination}
\input{Sections/5_discussion}

\input{Sections/6_conclusion}

\bibliography{main}
\bibliographystyle{tmlr}

\appendix

\end{document}

%% file: math_commands.tex
\usepackage{amsmath,amsfonts,bm}

\def\eqref#1{equation~\ref{#1}}

\def\1{\bm{1}}

\DeclareMathAlphabet{\mathsfit}{\encodingdefault}{\sfdefault}{m}{sl}
\SetMathAlphabet{\mathsfit}{bold}{\encodingdefault}{\sfdefault}{bx}{n}



%% file: Sections/1_introduction.tex
\section{Introduction}
\label{sec:intro}

In the rapidly evolving landscape of artificial intelligence (AI), large language models (LLMs) have emerged as a pivotal tool, driving innovation across a wide spectrum of applications, from natural language processing (NLP) and automated content creation \citep{achiam2023gpt,betker2023improving} to complex decision-making systems and autonomous agents \citep{wei2022chain,li2023chain,yang2023auto,wu2023autogen}. At their core, these models rely on extensive datasets \citep{commoncrawl,gao2020pile} to learn about language and the world, and generate responses that are increasingly indistinguishable from human-authored writing \citep{instructgpt}. However, the integrity of these datasets is paramount, as any \emph{contamination} can significantly impair the models' effectiveness, reliability, and generalization. 
\emph{Contamination} refers to leakage of evaluation data within the training data, an undesirable phenomenon which will inflate the model's performance on the evaluation data.

We identify several primary sources of such contamination that can arise during large-scale data collection and curation. First, unfiltered web scraping to collect pre-training data often introduces data with minimal oversight, increasing the likelihood of inadvertently capturing benchmark or test datasets that are publicly available online. Second, contamination may result from the unintentional inclusion of evaluation materials during preprocessing or aggregation of training corpora, particularly when datasets are not rigorously audited. Finally, the reuse of proprietary or copyrighted material without robust provenance tracking can also lead to inadvertent overlap between training and evaluation sets. These factors underscore the need for meticulous dataset design and transparent documentation practices to ensure the validity of experimental results.

Contamination poses a multifaceted challenge, threatening not only the technical accuracy of LLMs but also their ethical and commercial viability. In high-stakes scenarios where imprecision can have dramatic consequences, such as medical diagnosis, legal advice, or financial services, the repercussions of relying on contaminated data can be profound. Moreover, the allure of leveraging LLM outputs to attract investment underscores a pressing commercial dimension. As businesses increasingly integrate AI-driven insights into their strategic planning and operational decisions, the assurance of data purity becomes intertwined with potential market success and valuation. 
Lastly, in the current intense race to build the most powerful LLM \citep{touvron2023llama,gemini,jiang2024mixtral,bai2023qwen,bi2024deepseek,young2024yi,Gemma2024OpenModels,chen2023chatgpt}, the community is struggling to settle on a fixed subset of benchmarks as LLMs performance increases fast \citep{arc,hellaswag,mmlu,truthfulqa,winogrande,gsm8k}, an issue which is further fueled by underlying contamination. 
This landscape necessitates a comprehensive survey of contamination detection in LLMs.

\subsection{Contamination and Contamination Detection Types}

Contamination may span only part of a (x, y) data point where x is the input and y the corresponding label. If we consider whether the contaminated data contains only x or both x and y, contamination can be divided into \emph{input contamination} and \emph{input+label contamination}. 

Detecting contamination in a LLM is a broad issue that we divide into \emph{open-data} contamination detection and \emph{closed-data} contamination detection depending on the nature of the training set. On the one hand, \emph{open-data} refers to the scenario where the LLM pre-training data is known, enabling direct comparisons with the evaluation dataset. On the other hand, \emph{closed-data} contamination refers to the more and more prevalent use case of an unknown pre-training set. This latter case is more challenging and necessitates an examination of the LLM's behavior on evaluation data points. 

Detecting contamination is further complicated by varying level of access to the LLM. With \emph{white-box} access, the LLM and all weights are available, \eg\ a local model. In \emph{gray-box} access, model weights are not available to the researcher but output token probability distributions are. Most closed-source setups, \eg\ GPT-4, follow a \emph{black-box} setup where only the output text can be accessed from the API. We will review in this study contamination detection methods specifically tailored for each of these levels of model access.

\input{Figures/fig_1_classification}

\subsection{Contributions and Structure of this Survey}

To push forward research in contamination detection, we make the following contributions:
\begin{itemize}[leftmargin=*,itemsep=0.1em]
    \item To our knowledge, we are one of the first to thoroughly review the field of contamination detection in LLMs, unrestricted to a specific type of methods or a specific performance aspect. We review all reported contamination detection methods up until early 2025, totaling more than 50 detection techniques and more than 120 relevant papers. We followed an inclusive approached and kept all open-access papers relevant to the problem of contamination detection in LLMs, regardless of publication status. The vast majority of resulting papers which are included in this survey are from 2023-2025.

    \item We categorize contamination detection into two broad types of use cases vastly differing in the techniques involved, namely \emph{open-data} and \emph{closed-data} contamination detection, and review all existing work in each category. Our taxonomy further classifies detection techniques into finer-grained categories, each following their own assumptions and model requirements. 

    \item Through our detailed classification and review of the types of contamination detection, we not only contribute to the academic and practical understanding of data contamination issues but we also highlight the pressing need for strategies that mitigate these risks.
\end{itemize}

We refer to \Cref{fig:cls} for an overview of our proposed classification of contamination detection methods. 

The rest of this paper is structured as follows. Section 2 formally defines contamination and associated concepts. Section 3 explores methods and findings related to open-data contamination detection. Section 4 addresses the issue of closed-data contamination detection. Section 5 discusses the current and future challenges in the field, including best practices, and reviews all new datasets proposed to detect contamination.

\subsection{Comparison to Related Surveys}

There already exists surveys on the broad issue of contamination.

\citet{deng2024unveiling} survey contamination in language models, covering the problem from detection, to its impact and methods of remediating it. Compared to their work, our survey solely focuses on detecting contamination, and covers a more exhaustive set of contamination detection techniques. 

The work of \citet{palavalli2024taxonomy} proposes a taxonomy of the different types of contamination. In \Cref{fig:cls}, our second-level breakdown of closed-data contamination detection into \emph{dataset-level} methods and \emph{instance-level} methods echos their categories. Unlike their survey, we review contamination detection only, and we do so at large, unrestricted to a specific form of contamination such as a particular noising then injection of test data points into the training set. 

Unlike the two surveys above, \citet{fu2024does} focus on contamination detection, like our work. They also propose a taxonomy of detection methods, which has a few similarities with ours (like instance similarity, which we cover within open-data methods), but our classification is more fine-grained (four levels of classification, whereas their classification only has one). The authors review assumptions behind certain detection methods and challenge their true efficiency. Notably, they show that membership-inference attacks (MIAs) perform poorly in detecting pre-training contamination. Our study does not specifically focus on the success of contamination detection methods: we are rather interested in describing in depth the whole landscape of contamination detection, and explain how each method works, including requirements in terms of model access (\Cref{tab:model_access}) and best suited training stage (\Cref{tab:training_stage}).

%% file: Figures/fig_1_classification.tex
\begin{figure*}[t]
\scriptsize
    \centering
    \begin{adjustbox}{width=0.95\textwidth}
        \begin{forest}
        for tree={
                forked edges,
                grow'=0,
                draw,
                rounded corners,
                node options={align=center},
                text width=2.7cm,
                s sep=6pt,
                calign=edge midpoint, 
            },
                [Contamination \\Detection Methods, parent
                    [Open-Data \\Contamination Detection, fill=yellow!30, domain_box
                        [String \\Matching, fill=orange!30, domain_box
                            [Instance-Level \\String Matching, fill=red!20, domain_box
                                [{GPT-2 \citep{radford2019language}, GPT-3 \citep{brown2020language}, PaLM \citep{anil2023palm}, GPT-4 \citep{achiam2023gpt}, Llama-2 \citep{touvron2023llama}, Llama-3 \citep{dubey2024llama}, Qwen-2.5-Coder \citep{hui2024qwen2}, Longest-Match \citep{singh2024evaluation}}, models
                                ]
                            ]
                            [Dataset-Level \\String Matching, fill=red!20, domain_box
                                [{\citet{dodge2021documenting,li2023open,deng2023investigating}}, models
                                ]
                            ]
                        ]
                        [Embeddings \\Similarity, fill=orange!30, domain_box
                            [{Platypus \citep{lee2023platypus}, Phi-1 \citep{gunasekar2023textbooks}, \citet{riddell2024quantifying}}, models
                            ]
                        ]
                        [Paraphrase \\Detection, fill=orange!30, domain_box
                            [{LLM Decontaminator \citep{yang2023rethinking}}, models]
                        ]
                    ]
                    [Closed-Data \\Contamination Detection, fill=yellow!30, domain_box
                        [Dataset-level, fill=orange!30, domain_box
                            [Performance \\Analysis, fill=red!30, domain_box
                                [Train-Test Performance \\Gap, fill=purple!20, domain_box
                                    [{CAP \cite{zhao2024cap}}, models
                                    ]
                                ]
                                [Controlled Training \\Setup, fill=purple!20, domain_box
                                    [{\citet{jiang2024investigating,palavalli2024taxonomy}}, models
                                    ]
                                ]
                                [Performance Change Through Time, fill=purple!20, domain_box
                                    [{\citet{aiyappa-etal-2023-trust,roberts2023data,li2023task,cao2024concerned}}, models
                                    ]
                                ]
                            ]
                        ]
                        [Instance-level, fill=orange!30, domain_box
                            [Membership-Inference \\Attacks (MIA), fill=red!30, domain_box
                                [Reference-based \\MIA, fill=purple!20, domain_box
                                    [{RECALL \citep{xie2024recall}, MaxNorm\citep{meeus2024did}, MIA-Tuner \citep{fu2024mia}, PREMIA \citep{feng2024exposing}}, models
                                    ]
                                ]
                                [Reference-free \\MIA, fill=purple!20, domain_box
                                    [{SPV-MIA \citep{fu2023practical}, CAMIA \citep{chang2024context}, S2-MIA \citep{li2024generating}, EM-MIA \citep{kim2024detecting}}, models
                                    ]
                                ]
                            ]
                            [Model \\Memorization, fill=red!30, domain_box
                                [Quantifying \\Memorization, fill=purple!20, domain_box
                                    [{ Dynamic Soft Prompting \citep{wang2024unlock-memorizing},\citet{elangovan2021memorization,magar2022data,carlini2023quantifying,Gemma2024OpenModels}}, models
                                    ]
                                ]
                                [Prompt \\Engineering, fill=purple!20, domain_box
                                    [{\citet{nasr2023scalable}, \emph{According to} prompting \citep{weller2023according}, Guided prompting \citep{golchin2023data}, Local order quiz \citep{samuel2024towards}}, models
                                    ]
                                ]
                                [Cloze \\Tasks, fill=purple!20, domain_box
                                    [{TS-Guessing \citep{deng2023investigating}, \citet{chang-etal-2023-speak,ranaldi2024investigating}}, models
                                    ]
                                ]
                            ]
                            [Model \\Confidence, fill=red!30, domain_box
                                [Likelihood, fill=purple!20, domain_box
                                    [{Min-K\% Prob \citep{shi2024detecting}, CDD and TED \citep{dong2024generalization}, DC-PDD \citep{zhang2024pretraining}, Min-K\%++ \citep{zhang2024min}, SURP \citep{zhang2024adaptive}, \citet{li2023estimating,oren2023proving,xu2024benchmarking}}, models
                                    ]
                                ]
                                [Probing Hidden Layers, fill=purple!20, domain_box
                                    [{DICE \citep{tu2024dice}, \citet{liu2024probing}}, models
                                    ]
                                ]
                            ]
                        ]
                    ]
                ] 
        \end{forest}
    \end{adjustbox} 
    \caption{\small Classification of contamination detection methods reviewed in this paper.}
    \vspace{-2em}
    \label{fig:cls}
\end{figure*}

%% file: Sections/2_definitions.tex
\section{Formal Definitions}
\label{sec:definitions}

Formally, let $M$ be a model (e.g. Llama-3 \citep{dubey2024llama}) and $\mathcal{D}$ be a data distribution on an input space $\mathcal{X}$. We denote by $D_{M}$ the dataset that $M$ was pre-trained on (e.g, The Pile \citep{gao2020pile}), and $D_{\text{eval}}$ an evaluation dataset of interest (e.g, MMLU \citep{mmlu}).

\begin{definition}[Contamination]
We say that the dataset $D_{\text{eval}}$ is \emph{\textbf{contaminated}} by the dataset $D_{M}$  if: 

\begin{equation}
    \exists x \in D_{\text{eval}}, \exists x' \in  D_{M}, f(x, x') >= \tau
    \label{contamination}
\end{equation}

where $f$ is a similarity function, and $\tau$ a similarity threshold. $\tau=1$ corresponds to \textbf{verbatim} or \textbf{strict} contamination, but we do not restrict ourselves to this setup in this paper. 
\end{definition}

Contamination signs often include inflated performance on evaluation benchmarks, and model memorization of entire substrings. 
\textbf{Contamination \emph{detection}} refers to techniques assessing whether \Cref{contamination} holds true or not.

\begin{definition}[Membership Inference Attack]
A \textbf{membership inference attack (MIA)} is a function attempting to assess whether a data point was part (a member of) the training set $D_{M}$, and can be noted as:

\begin{equation}
    A: \mathcal{X} \times M(\mathcal{X}) \rightarrow \{0,1\}
\end{equation}

where 

\begin{align*}
A(x, M(x)) &= 1 \quad \text{if } x \in D_{M} \\
A(x, M(x)) &= 0 \quad \text{otherwise}
\end{align*}
\end{definition}

Both concepts of contamination detection and membership inference are closely related yet fundamentally different: a membership inference attack can be used to detect contamination, but membership inference attacks can also be defined and succeed without the existence of contamination.

\begin{definition}[Model Memorization]
\label{def:memorization}
Let us note $G(p)$ the model's natural language output on natural language prompt $p$. We say that $M$ has \textbf{memorized} a data point $x \in D_{M}$ if:

\begin{equation}
\Pr[G(p_{x}) = x \mid x \in D_{M}] \gg \Pr[G(p_{x}) = x \mid x \sim \mathcal{D}]
\end{equation}
for some prompt $p_{x}$ that elicits $x$ (typically, $p_{x}$ is a truncation of $x$).

\end{definition}

In other words, memorization refers to the LLM's capability to recreate a training data verbatim when prompted accordingly.

\color{black}

%% file: Sections/3_open_data_contamination.tex
\section{\emph{Open}-Data Contamination Detection}
\label{sec:open}

In the open-data setup, we have access to the (model-specific) pre-training set $D_{M}$. 
In this case, contamination detection falls back to investigating the intersection $D_{M} \cap D_{\text{eval}}$ for any evaluation dataset $D_{\text{eval}}$.
In the following, we review existing open-data contamination detection methods, which directly compare ensembles of data points.

\subsection{String Matching}

The most straightforward method to compute overlap between textual datasets is string matching or analyzing the set of words intersecting two texts. 

\subsubsection{Instance-Level String Matching}

Several LLM pre-training reports assess contamination levels by using string matching techniques. 
GPT-2 \citep{radford2019language} calculates contamination as the percentage of 8-grams from a particular evaluation set that are also found in the WebText training set. An average of 3.2\% overlap between common LM datasets’ test sets and the WebText dataset used for pre-training is discovered, raising the first concerns that LLM performance may be affected by memorization of pre-training data. 
GPT-3~\citep{brown2020language} scans data points having a 13-gram collision with anything in the pre-training Common Crawl (C4) dataset~\citep{commoncrawl}. Some earlier benchmarks sourced from Wikipedia, the Children’s Book Test dataset~\citep{hill2015goldilocks}, Quac \citep{quac}  and SQuAD 2.0 \citep{squad2}, are nearly entirely contaminated.
PaLM~\citep{chowdhery2023palm} identifies 10 evaluation datasets at risk of contamination per their construction process, and for each dataset, partitions it into a clean and a contaminated subset based on whether at least 70\% of the 8-grams of the data point can be found at least once in the training set. Both GPT-3 and PaLM authors conclude that the performance gap between clean sets and contaminated sets is mostly small across sets of various contamination degrees.
GPT-4~\citep{achiam2023gpt} measures cross-contamination between evaluation sets and pre-training data by computing 50-characters substring collisions. After examining GPT-4 performance gap between contaminated and clean subsets, the authors conclude that contamination overall has very little effect on the reported zero-shot results.
Llama-2 \citep{touvron2023llama} refines the methods above by using \emph{tokens} instead of \emph{words} to measure contamination, and asserts that a token is contaminated if it appears in any token n-gram longer than 10 tokens in both the evaluation sample and the training set. The formatting of prompts used for actual evaluation is also considered as it affects the performance more significantly. From there, the contamination level of a sample is defined as its percentage of contaminated tokens. The authors also introduce the usage of a \emph{skip-gram} budget allowing matched spans between the evaluation sample and the training set to differ in at most four positions.
Llama-3 \citep{dubey2024llama} follows a similar technique to Llama-2, but using 8-gram tokens instead of 10-gram. The ratio of contaminated tokens is varied to find the value with highest performance gain across models. Authors conclude that performance on PiQA \citep{bisk2020piqa} and HellaSwag \citep{hellaswag} is affected by contamination. 
Recently, Qwen-2.5-Coder \citep{hui2024qwen2} removes all training data points with a 10-gram collision with the test set. Popular benchmarks HumanEval \citep{chen2021evaluating}, MBPP \citep{mbpp}, GSM8K \citep{gsm8k}, and MATH \citep{math} are affected. 

\Cref{tab:string_matching} summarizes contaminated datasets found through all string-matching contamination detection described above. Notably, PIQA \citep{bisk2020piqa}, Winograd \citep{winograd}, HumanEval \citep{chen2021evaluating} and HellaSwag \citep{hellaswag} are flagged by at least two contamination detection techniques.

\input{Tables/string_matching_results}

\citet{singh2024evaluation} review several string-matching contamination detection techniques, including the ones used by \citet{brown2020language} (Token-Match), \citet{chowdhery2023palm} (Ngram-Match) and \citet{touvron2023llama} (Token-Extend). They introduce a new method called Longest-Match which measures the fraction of tokens part of the longest contaminated token span, avoiding the issue of frequent token spans artificially inflating contamination levels with other methods. Similarly to \citep{touvron2023llama}, Longest-Match also allows for skipgram. Longest-Match finds the highest Estimated Performance Gain (EPG) between contaminated and clean evaluation subsets. Through Llama-1 \citep{llama1} and Pythia \citep{biderman2023pythia} series of LLMs, the Llama pre-training corpus and The Pile \citep{gao2020pile} are found to be contaminated for many popular evaluation benchmarks, notably BigBench \citep{bigbench}, HellaSwag \citep{hellaswag}, PiQA \citep{bisk2020piqa}, MMLU \citep{mmlu} and TriviaQA \citep{joshi2017triviaqa}. 

\subsubsection{Dataset-Level String Matching}

Other studies analyze overlap between the training split and the evaluation or test split of the \emph{same} dataset to assess contamination levels. 
While standard practice in machine learning assumes that training, validation and test splits of the same dataset are disjoint sets sampled from the same distribution, in practice, for very large datasets it may happen that the training set still overlaps with the other splits.
\citet{dodge2021documenting} investigate the C4~\citep{commoncrawl} dataset and study benchmark data contamination, which measures how much training or test datasets from downstream NLP tasks appear in the C4 pre-training corpus. After studying both input-label and input contamination, they discover varied contamination, ranging from less than 2\% to over 50\%.
\citet{li2023open} compute the METEOR~\citep{banerjee-lavie-2005-meteor} score between matched pages from CommonCrawl and queries from Bing API and consider those with scores over 0.75 as contaminated. Detected contamination levels range from 1\% on Winogrande \citep{winogrande} to 47\% on C-Eval \citep{huang2023c}.
\citet{deng2023investigating} propose to detect contamination by retrieving top-10 documents from pre-training datasets The Pile \citep{gao2020pile} or C4 \citep{commoncrawl}, then splitting retrieved documents into 13-gram chunks, and computing overlap metrics between these chunks and the evaluation data point's chunks. TruthfulQA \citep{truthfulqa} exhibits high overlap with the pre-training datasets.

\subsection{Embeddings Similarity}

Beyond simple string matching, computing cosine similarity between embeddings offers an attractive alternative, as this process is more robust to surface-form vocabulary changes like paraphrases.

\citet{lee2023platypus} prevent contamination in their Open-Platypus dataset by removing test questions which have a cosine similarity (using SentenceTransformer embeddings \citep{reimers2019sentence}) greater than 80\% against any training item. 
Phi-1~\citep{gunasekar2023textbooks} runs a data contamination analysis between their CodeExercises dataset and the evaluation set HumanEval \citep{chen2021evaluating}. They show that embeddings-based retrieval between code snippets using L2 distance between normalized CodeGen-Mono 350M \citep{nijkamp2022codegen} embeddings is effective, whereas n-gram overlap fails in the coding domain as it cannot capture the similarities in the logic between two coding functions.

The work from \citet{riddell2024quantifying} provides a hybrid approach combining string matching and embeddings similarity: an \emph{aggregated score} is designed as the maximum between the Levenshtein edit distance and the similarity from the Dolos toolkit \citep{maertens2022dolos}. Widely used code generation benchmarks MBPP \citep{mbpp} and HumanEval \citep{chen2021evaluating} are contaminated with pre-training sets The Pile \citep{gao2020pile} and especially The Stack \citep{kocetkov2022stack}, which significantly inflates performance of models like CodeGen-NL \citep{nijkamp2022codegen}.

\subsection{Paraphrase Detection}
 
Embeddings similarity is limited by its need to choose an appropriate similarity threshold. To improve robustness, some methods attempt to use LLMs themselves to detect contamination in LLMs, leveraging their high performance in zero-shot paraphrase detection \citep{witteveen2019paraphrasing,abaskohi2023lm}.

\citet{yang2023rethinking} argue that contamination checks become challenging because of the presence of “rephrased samples”, which have the same semantics but different surface form as the original sample. Therefore, they first use embeddings similarity search to get the top-k similar samples with a given test sample, then prompt a strong LLM (namely, GPT-4) to examine whether any of the top-k samples is too close to the test case. Results show that this \emph{LLM Decontaminator} algorithm works significantly better than existing methods at tagging contaminated samples.

%% file: Tables/string_matching_results.tex
\begin{table}[]
\resizebox{0.98\columnwidth}{!}{
\begin{tabular}{lll}

\toprule 

\textbf{LLM}   & \textbf{String Matching} & \textbf{Contaminated Datasets} \\

\midrule 

GPT-2 \citep{radford2019language}   & 8-gram words           & CoQA \citep{coqa}, LAMBADA \citep{lambada} \\
GPT-3 \citep{brown2020language}     & 13-gram words          & PIQA \citep{bisk2020piqa}, Winograd \citep{winograd}, Children’s Book Test \citep{hill2015goldilocks}, WikiText-2 \citep{merity2016pointer}, \\ 
& & WikiText-103 \citep{merity2016pointer}, enwik8, text8 \\
PaLM \citep{chowdhery2023palm}      & 8-gram words           & Winograd \citep{winograd}, SQuAD-2 \citep{squad2}, WSC, ReCoRD, CB \\
GPT-4 \citep{achiam2023gpt}         & 50-gram characters     & HumanEval \citep{chen2021evaluating} \\
Llama-2 \citep{touvron2023llama}    & 10-gram tokens         & HellaSwag \citep{hellaswag}, MMLU \citep{mmlu} \\
Llama-3 \citep{dubey2024llama}      & 8-gram tokens          & PiQA \citep{bisk2020piqa}, HellaSwag \citep{hellaswag}, AGIEval \citep{agieval}, BIG-Bench Hard \citep{bigbench}, \\
& & BoolQ \citep{boolq}, OpenBookQA \citep{openbookqa}, QuaC \citep{quac}, SiQA \citep{sap2019socialiqa} \\
Qwen-2.5-Coder \citep{hui2024qwen2} & 10-gram words          & HumanEval \citep{chen2021evaluating}, MBPP \citep{mbpp}, GSM8K \citep{gsm8k}, MATH \citep{math} \\ 

\bottomrule
\end{tabular}
}
\caption{\small Datasets flagged with a high-level of contamination for several LLM pre-training studies, assessing contamination through string matching between the pre-training corpus and the evaluation samples.}
\label{tab:string_matching}
\end{table}

%% file: Sections/4_closed_data_contamination.tex
\section{\emph{Closed}-Data Contamination Detection}
\label{sec:closed}

In closed-data contamination detection, the pre-training set $D_{M}$ is unknown. Since $D_{M}$ is not accessible, researchers have to use tools analyzing the model's behavior on the evaluation dataset $D_{\text{eval}}$.

There are two broad ways to assess closed-data contamination: examining model behavior aggregated on the whole dataset, and focusing on instance-level outputs. 
For the former, the main method, which we explore in \Cref{subsec:perf}, consists in tracking model performance on several datasets, sampled from different distributions, notably through time. 
For the latter, we review methods ranging from the strongest assessment of contamination to the weakest: membership-inference attacks, which assess the presence of an entire data point in the pre-training set (\Cref{subsec:mia}) ; then memorization of spans of text by the model (\Cref{subsec:mem}) ; and lastly methods analyzing model confidence on specific tokens (\Cref{subsec:conf}).

\subsection{Performance Analysis}
\label{subsec:perf}

We first review a line of research assessing model contamination through simple performance analysis. $M$ is applied to different evaluation datasets, and patterns in evaluation scores may indicate contamination.

\subsubsection{Train-Test Performance Gap}

On a dataset where the training set and the test set are sampled from the same distribution, a machine learning system is expected to perform on the test set slightly worse, or at best similar to, the training set. \citet{zhao2024cap} leverage this fact to distinguish between standard fine-tuning and contamination, focusing on domain-specific setups such as financial data. They introduce a Performance Consistency Ratio (PCR), equal to the LLM performance on the test set divided by its consistency, where the latter measures the LLM's ability to produce the same output across several reasoning paths. A PCR on the training set significantly greater than on the test set is typical of fine-tuning ; the opposite case points out at contamination. Findings include potential contamination of Baichuan-13B \citep{yang2023baichuan} on FinEval \citep{zhang2023fineval} and of the FinMA model \citep{xie2023pixiu} on the FinQA dataset \citep{chen2021finqa}.

\subsubsection{Controlled Training Setup}

\citet{jiang2024investigating} delve deep into the impact of evaluation data leakage within the pre-training set by running a series of controlled pre-training experiments with GPT-2 \citep{radford2019language}. They pre-train from scratch GPT-2 in three different ways: with the original pre-training set, with additional input texts from four evaluation datasets (SST-2, MMLU \citep{mmlu}, CNN/DM \citep{cnndm} and SQuAD-v1 \citep{squad}), and with input texts and labels from these same evaluation datasets. Contaminated pre-training clearly boosts performance on all downstream tasks. This was also shown by the work of \citet{palavalli2024taxonomy} in their Verbatim setup when pre-training GPT-2. Findings also reveal a U-shaped relationship between the number of times that contaminated data points are included in the pre-training, and model performance. Authors also show that the n-gram contamination detection method from Llama-2 \cite{touvron2023llama} is not effective at detecting contaminated data points, calling for more robust methods.

\subsubsection{Performance Change Through Time}

The most straightforward performance analysis consists in plotting performance against \textbf{\emph{time}}, which leverages the fact that evaluation datasets released after a LLM is pre-trained are new and should not be contaminated. 

\citet{aiyappa-etal-2023-trust} are among the first to question the non contamination of ChatGPT training data by popular evaluation benchmarks. Their run of the January 30th, 2023 version of ChatGPT on the SemEval 2016 dataset shows a 12.5 macro-F1 score improvement compared to a prior experiment with the v1 of ChatGPT, released two months prior in November 2022.
In the code domain, \citet{roberts2023data} investigate the performance of GPT-3.5 and GPT-4 on datasets released before and after the training cutoff for both models. The analysis underscores a statistically significant positive correlation for problems catalogued on GitHub prior to the GPT models' training cut-off, suggesting that their inclusion in the training data enhanced their performance on related tasks. Concurrent to their work, \citet{li2023task} also plot the performance of several LLMs including GPT-3 \citep{brown2020language} and open-source LLMs on general natural language understanding tasks against time, specifically before and after each LLM's training data cutoff date. They find that LLMs are much more likely to improve on a trivial majority baseline for datasets released before their cutoff date, suggesting some contamination. While evaluating on fresh data guarantees a contamination-free performance evaluation, it is hard to scale this method in time as it requires frequent update of evaluation sets. Besides, the work of \citet{cao2024concerned} shows that evaluating on data posterior to the LLM training cutoff time may not always show a decrease in performance: in the code domain, the opposite may happen, complicating the usage of training cutoffs as a contamination detection method.

\subsection{Membership-Inference Attacks}
\label{subsec:mia}

Extracting knowledge from a language model has traditionally been addressed in the field of privacy-preserving machine learning. Adversarial attacks were crafted as \textbf{membership inference attacks (MIA)} to access training data points memorized by a black-box language model, potentially compromising sensitive private information \citep{carlini2021extracting,carlini2022membership}. If there is access to data points which are known not to be present in the training set, membership inference attach is referred to as \textbf{reference-based MIA} ; other cases belong to the setup of \textbf{reference-free MIA}.

\subsubsection{Reference-based Membership-Inference Attacks}

\citet{xie2024recall} derive a MIA named RECALL based on the conditional language capacity of LLMs. RECALL conditions with non-member prefixes (for instance, sampled after pre-training cutoff date): log-likelihood drops more on average for member data points than for non-members, yielding a simple yet powerful MIA. The method reaches state-of-the-art ROC-AUC on the WikiMIA dataset \citep{shi2024detecting}.
\citet{meeus2024did} revisit MIA in the era of LLMs and define document-level MIA, which is the task of detecting whether an entire document (book, scientific paper, etc) has been trained on. A reference set is built by sampling documents known to be part of the pre-training set (e.g. Common Crawl), and data posterior to the pre-training cutoff date, which is guaranteed to be uncontaminated. Then, the authors derive features from predicted token probabilities by the LLM, which are normalized in several fashions. A meta-classifier predicts membership inference, reaching ROC-AUC as high as 0.86. 

In the last stage of LLM post-training, PPO \citep{schulman2017proximal} or DPO \citep{rafailov2023direct} is usually employed to train LLM on preference data. \citet{feng2024exposing} show that LLMs are vulnerable to MIA on the preference data, especially when using DPO.
LLMs themselves can be instruction-tuned to detect if a data point is a member of their pre-training data. MIA-Tuner \citep{fu2024mia} fine-tunes a soft prompt \citep{lester2021power} on several target open-source LLMs to classify membership inference, and outperforms all other methods on WikiMIA. Even though at inference time, non-members are not needed, the methodology should still be assessed as a reference-based MIA due to the usage of non-members during the LLM fine-tuning.

\subsubsection{Reference-free Membership-Inference Attacks}

\citet{fu2023practical} argue that both reference-based and reference-free MIA are unsatisfying with LLMs in practical scenarios. The latter rely on overfitting to assess membership inference. In the reference-free setup, they propose to prompt the target LLM to generate pseudo reference data. Semantically similar and semantically different paraphrases of target records are generated, and a membership signal is designed by measuring and comparing probabilistic variations between the target and reference models. This method reaches a ROC-AUC of 0.95 with Llama on the language modelling dataset WikiText-103.
\citet{chang2024context} analyze the dynamics of next-token prediction loss for open-source Pythia and GPT-Neo LLM series. They observe that for members, the loss tends to be smoother and to decrease as context increases, whereas non-members loss curves show a flat slope as well as quite frequent outlier values. From this observation the authors derive a state-of-the-art MIA technique called CAMIA.
EM-MIA \citep{kim2024detecting} introduces a MIA method iteratively refining the membership score used to assess MIA and the prefix score through an expectation-maximization algorithm. The prefix score measures how discriminative the data point is in classifying members and non-members when used as a prefix. EM-MIA reaches state-of-the-art ROC-AUC scores on WikiMIA, however on a custom benchmark where members and non-members distributions are almost identical, performance drops back to random range, as for all other reported MIA methods.

\subsubsection{Limitations of Membership-Inference Attacks}

Membership-Inference attacks are hard in practice with neural networks. \citet{duan2024membership} evaluate a series of MIAs on the Pythia LLMs (from 70M to 12B parameters), pre-trained on the Pile. Across types of attacks and domains, performance hardly surpasses random guessing (ROC-AUC inferior to 0.6). This failure is attributed to the common practice in LLM development of pre-training only for a single epoch and on a too large dataset, and the fact that the border between members and non-members is not clear. By splitting both groups based on time, where non-members are drawn from datasets posterior to the pre-training cutoff date, MIA accuracy drastically improves. This finding motivates the monitoring of LLMs performance through time described above. \citet{meeus2024sok} echo these findings from \citet{duan2024membership}: they train a trivial bag-of-words model on MIA datasets constructed by splitting on time and reach a high ROC-AUC. Authors argue that proper MIA evaluation should be done on a randomized train-test split. Concurrent work from \citet{maini2025llm} also question the validity of membership inference attacks on a time-split of Wikipedia. An experiment replacing member sentences by sentences from the validation set of The Pile \citep{gao2020pile} using Pythia models \citep{biderman2023pythia} shows membership inference tests still reaching an ROC-AUC of 0.7 ; proving that they merely capture a temporal distribution shift rather than actual membership to the pre-training set. 
Despite this lack of success of MIA on pre-training sets of LLMS, \citet{li2024generating} show that similarity between the target sample and the generated output of a LLM-powered RAG system yields a powerful MIA on the RAG database. On Natural Questions \citep{kwiatkowski2019natural} and TriviaQA \citep{joshi2017triviaqa}, the method reaches more than 0.85 in both ROC-AUC and PR-AUC.

\subsection{Model Memorization}
\label{subsec:mem}

An intuitive approach to detect contamination in LLMs is to frame it as a \textbf{memorization} detection problem.
While MIAs assess the presence of an \emph{entire} data point into the training set, here memorization is more granular and detects memorized sequences of tokens which may not be full data points (documents), but rather subsets of them. The rate of memorized data points can be used to assess the presence of contamination. For this endeavor, it is important to first be able to quantify memorization itself.

\subsubsection{Quantifying Memorization}

\citet{elangovan2021memorization} highlight that some LLMs may benefit from an overlap between training and test sets, and merely memorize data points in this intersection, inflating their test set performance. This overlap is due to the common practice of simple random data shuffling in NLP tasks. By employing a bag-of-words approach to assess text similarity, the study provides a foundational method for identifying and mitigating leakage, though it acknowledges the complexity of semantic similarity measurement and the potential for more sophisticated methodologies to refine overlap detection.
\citet{magar2022data} find evidence of contamination across several benchmarks, impacting the performance of models like GPT-3. They propose to quantify the effect of contamination by training on a mix of general and task-specific data, then comparing performance on seen versus unseen instances to measure memorization and exploitation. This methodology highlights the nuanced relationship between memorization and exploitation.
\citet{carlini2023quantifying} propose a simple method to detect memorization by measuring the completion rates of randomly sampled prefixes from the training set, and discuss the relations between memorization with model capacity, data repetition, context size. Memorization is more widespread than previously thought, and model size, data frequency and prompt length are factors leading to an increase in memorization.

The team from Google developing the Gemma model series \citep{Gemma2024OpenModels} emphasizes the potential vulnerabilities of aligned models to adversarial attacks inducing recitation of memorized training data. Through comparable methods to \citet{anil2023palm}, they analyze \emph{discoverable} memorization across 10k documents from the Gemma pre-trained models' corpora. The analysis distinguishes between \emph{exact} memorization — where the model's output precisely matches the source — and \emph{approximate} memorization, gauged by a 10\% edit distance threshold. Gemma models exhibit memorization rates comparable to those of the PaLM and PaLM-2 models. 

As we have seen in \Cref{def:memorization}, detecting memorization sums to finding a (prefix, suffix) mapping such that the corresponding suffix can be perfectly completed when conditioning the model on the prefix. For example, \citet{mustafa2023memorized-prompt-tuning} append a learnable soft prompt penalized through \emph{gradient ascent} to prevent memorizing the original suffix. \citet{wang2024unlock-memorizing} employs a transformer-based generator to approximate such mapping of prefixes, and use it to estimate the memorization rate on unseen corpus.

\subsubsection{Prompt Engineering}

Recently, researchers have engineered elaborated prompting techniques that elicit data point completion by the LLM, where an output suspiciously close to the actual training data point indicates memorization, which in turn means contamination. Standard prompting asking the LLM to complete a data point may fail due to the LLM's alignment during RLHF \citep{ouyang2022training}.

By prompting with an off-(training)-distribution input, \citet{nasr2023scalable} show that ChatGPT can regurgitate entire chunks of training data, including sensitive personal-level information. The study illustrates that even models trained with alignment techniques, aimed at reducing the emission of memorized data, can be induced to divulge substantial amounts of sensitive information, calling for the development of more robust defenses. The implications of this research are far-reaching, emphasizing the importance of addressing the privacy vulnerabilities inherent in the deployment of LLMs.
\citet{weller2023according} highlight the potential for steering LLMs towards generating more factual and grounded content through effective prompting strategies. 
By appending instructional phrases such as \emph{Accoding to} that encourage quoting from specific corpora, this study observes improvements in grounding as measured by the Quoted Information Precision (QUIP) Score. This method shows promise across various domains and corpora, indicating its versatility and effectiveness in leveraging LLMs' memorized knowledge for generating more accurate and reliable responses.

\citet{golchin2023time} make use of GPT-4's abilities in \emph{guided prompting}, an enhanced prompting process where the completion prompt includes extra information such as the dataset name. Contamination is then assessed based on the averaged difference in performance between standard and guided prompting, or if GPT-4 with in-context learning finds an exact match or two near-exact matches in the guided completions. This latter method with GPT-4 is highly accurate in identifying contamination (92\%–100\% accuracy). Moreover, the investigation highlights the prevalence of contamination in datasets such as AG News \citep{agnews}, WNLI \citep{winograd}, and XSum \citep{xsum}, emphasizing the critical need to address data integrity in LLM applications.
The same authors follow up with the Data Contamination Quiz (DCQ) evaluation framework for contamination detection in black-box LLMs \citep{golchin2023data}. LLMs are prompted with five completion options, where one option is the ground-truth text from the original dataset, three other options are paraphrases from GPT-4, and the last choice is None. The assumption is that if the LLM picks the exact answer, it is doing so out of memorization, and the authors show that this DCQ framework finds more contamination cases than the guided prompting method.
\citet{samuel2024towards} also propose a quiz framework to detect contamination. Their Local Order Quiz method consists in prompting the LLM with a data point and asking it to select the subsequent data point in the original data set, among four possibilities shown in the prompt (the correct next data point, and three other random ones). A higher accuracy that the baseline 25\% chance indicates that the LLM has memorized the data set. In practice, the IMDB \citep{maas2011learning} dataset shows high signs of contamination (subsequent data point selection accuracy over 80\%) for all three LLMs under study: GPT-4 \citep{achiam2023gpt}, Claude-3 \footnote{\url{https://www.anthropic.com/news/claude-3-family}} and Llama-3 \citep{dubey2024llama}.

\subsubsection{Cloze Tasks}

Cloze tasks constitute a special type of prompting where parts of the input is masked. 
\citet{chang-etal-2023-speak} design a cloze task in which the LLM is prompted with a book's passage, from which the character name has been replaced by a [MASK] token. Correctly predicting the character's name out of several passages per book assesses whether the LLM has seen this book during pre-training. The name cloze accuracy is very high for GPT-4 \citep{achiam2023gpt} on several popular books such as \emph{Alice's Adventures in Wonderland}, and shows a striking difference with BERT's \citep{devlin2019bert} performance on the same task. 
\citet{deng2023investigating} present the TS-Guessing method: one of the options in question-answering benchmarks test sets is hidden, and the LLM is asked to predict it. Exact match rates above 50\% for OpenAI models suggest contamination on the MMLU dataset \citep{mmlu}.
The work of \citet{ranaldi2024investigating} also uses a cloze task: they prompt the LLM to reconstruct columns names in SQL dumps, and compare the performance on an older and more recent dataset to assess contamination. GPT-3.5 accuracy falls from 33.42\% to 13.21\%, indicating contamination on the older Spider dataset \citep{yu2018spider}.

\subsection{Model Confidence}
\label{subsec:conf}

Going beyond simple recitation of memorized token sequences, researchers may now analyze the LLM's \emph{confidence} to detect contamination: a (too) high confidence on some tokens may signify prior exposure to the data. 
Confidence analysis yields several powerful contamination detection techniques, which we highlight require gray-box to white-box access to the LLM, as researchers need to obtain the output probability distribution for each token during inference.

\subsubsection{Likelihood}

Next-token likelihood and perplexity (by extension to a whole dataset) are widely used in contamination detection. 

Given a test tokens sequence $X$, the Min-K\% Prob technique \citep{shi2024detecting} consists in running the LLM through all tokens of $X$, and keeping track of the K\% tokens with the smallest predicted probability. Then, the average between such bottom probabilities is computed and $X$ is deemed contaminated if this average is \emph{too high}. 
In mathematical terms, noting $(x_{1}, \ldots, x_{n})$ the sequence of evaluation tokens under study, Min-K\% Prob is computed as :

\begin{equation}
    \text{Min-K\% Prob}(x) = \frac{1}{|\text{Min-K\%}(x)|}\sum\limits_{x_{i} \in \text{Min-K\%}(x)} \text{log}(p(x_{i}|x_{<i}))
\end{equation}

A significant drawback is the choice of value for the hyper-parameter K, which authors recommend to set to 20 as default. On WikiMIA, Min-K\% Prob reaches an average ROC-AUC of 0.72, and is the highest for every LLM tested. Notably, it outperforms other methods such as perplexity or the zlib compression entropy membership-inference attack. 
Due to its simplicity and effectiveness, Min-K\% Prob is gaining traction as one of the most popular contamination detection methods. \citet{zhu2024inference} use Min-K\% Prob as a first contamination detection step to identify leaked test samples before re-writing them, in order to build cleaner versions of popular evaluation datasets GSM8K \citep{gsm8k} and MMLU \citep{mmlu}.
Min-K\% Prob performance is very high ; yet the method has a major limitation: it does not account for the token frequency distribution. LLMs put a higher probability on more frequent tokens, thus a non-contaminated sequence made of only frequent tokens might get assigned a score higher than a sequence from the training set. To fix this, DC-PDD \citep{zhang2024pretraining} calibrates token probabilities and computes the cross-entropy between the token probability distribution and the token frequency distribution. The latter is estimated through a large-scale publicly available reference corpus. The tokens sequence is scored by averaging these token-level cross-entropy scores over the first occurence of each token. Contamination is assessed when this final score is above a pre-defined threshold. 
Min-K\%++ \citep{zhang2024min} proposes another extension of the idea of Min-K\% Prob: rather than simply getting next-token predicted probabilities, the authors propose a score which subtracts the expected log-probability and divides the score by the variance; in other words, normalizing the initial log-probability. The sequence scoring mechanism is then identical to the one of Min-K\% Prob. Formally, noting the expected probability over the next token $\mu_{x<i} = \mathbb{E}_{z \sim p(.|x_{<i})}[\text{log}(p(z|x_{<i}))]$ and its corresponding standard deviation $\sigma_{x<i} = \sqrt{\mathbb{E}_{z \sim p(.|x_{<i})}[(\text{log}(p(z|x_{<i})) - \mu_{x<i})^{2}]}$:

\begin{equation}
    \text{Min-K\%++}(x) = \frac{1}{|\text{Min-K\%}(x)|}\sum\limits_{x_{i} \in \text{Min-K\%}(x)} \frac{\text{log}(p(x_{i}|x_{<i})) - \mu_{x<i}}{\sigma_{x<i}}
\end{equation}

The motivation of the score built by Min-K\%++ is to assess whether an input form a mode, after observing that because of maximum likelihood training, training samples often become local maxima in the modeled distribution along each input dimension (here in the context of LLM, dimensions are tokens). Min-K\%++ reaches state-of-the-art on WikiMIA, outperforming Min-K\% Prob by up to 10 points.

\citet{li2023estimating} also work on token-level probabilities and compare perplexity on benchmark samples against memorized and clean baselines. The study finds significant memorization in recent models on popular reading comprehension and summarization benchmarks, while multiple-choice benchmarks show less evidence of contamination. This method provides a tool for the community to conduct rigorous contamination analysis, enabling more accurate and reliable model evaluation.
\citet{dong2024generalization} propose two novel likelihood-based contamination detection methodologies: CDD (Contamination Detection via output Distribution) and TED (Trustworthy Evaluation via output Distribution). CDD detects data contamination by observing the peakedness in the LLM's output distribution in a black-box manner. It represents a significant improvement over existing approaches, offering average relative improvements of 21.8\%-30.2\% in terms of Accuracy, F1 Score, and AUC metrics. TED corrects the LLM's output distribution to mitigate the impact of data contamination on evaluation metrics, significantly reducing performance improvements attributed to data contamination across various scenarios and contamination degrees.
The work of \citet{xu2024benchmarking} proposes to use perplexity and accuracy on next n-gram prediction. They compute the relative loss in performance on these metrics on paraphrased versions of the dataset created by ChatGPT compared to the original dataset. The difference of relative loss between training and test set is later used to assess contamination or \emph{benchmark leakage}. Experiments on GSM8K \citep{gsm8k} hint at contamination for Qwen \citep{bai2023qwen}, Aquila and InternLM \citep{cai2024internlm2} series of LLMs. 
\citet{zhang2024adaptive} leverage the concept of \emph{surprising tokens}: these are tokens where the LLM is very confident yet wrong in its prediction of the next token. The authors define a score named SURP which is the average predicted log-probability on the ground-truth token for tokens of the sequence where the model shows low entropy (and therefore, high confidence) and where this log-probability is relatively smaller. A higher SURP score is thresholded to assess contamination. The method slightly outperforms Min-K\% Prob on WikiMIA.

\citet{oren2023proving} take an original approach to analyze the LLM's likelihood on an evaluation dataset $D_{E}$. They run inference on $D_{E}$ and on shuffled versions of $D_{E}$. LLM log probabilities being statistically different on the non-shuffled dataset compared to the shuffled versions signifies contamination. The method relies on the assumption that if they are included in the pre-training set, evaluation datasets tend to be present with the same default ordering. We point out that detecting the LLM's memory of a dataset canonical order is an idea also explored by \citet{samuel2024towards} for a different contamination detection method.

\subsubsection{Probing Hidden Layers}

Through the use case of fine-tuning LLMs on mathematical reasoning datasets like GSM8K \citep{gsm8k}, \citet{tu2024dice} introduce a new paradigm in (closed-data) contamination detection. Their DICE approach consists in finding the layer of the fine-tuned LLM furthest away from its counterpart layer in the not fine-tuned same LLM. Then, a MLP binary classifier is inferred on the isolated layer's weights to predict contamination. This method is particularly effective for \emph{in-distribution} contamination detection, a setup defined by the authors as fine-tuning on a dataset from the same distribution as the evaluation dataset. In this setup, DICE reaches near-perfect ROC-AUC, outperforming all other methods, including notably Min-K\% Prob. 

\input{Tables/closed_data_model_access}

\citet{liu2024probing} also make use of the intermediate layers of a fine-tuned LLM. Their method first gathers a dataset of non-members and members ; where the former are pulled from after the pre-training cutoff or synthetically generated from ChatGPT. Then, the LLM is fine-tuned to classify membership between these two groups. Activations of intermediate layers of the fine-tuned LLM on both members and non-members are used as input of a probe classifier assessing membership. At inference time, the whole system feeds a target text to the fine-tuned LLM, extract its activations and feeds them to the probe classifier to check membership. The best layer to extract activation from is determined on a validation set. Results show greater ROC-AUC on WikiMIA \citep{shi2024detecting} and ArxivMIA \citep{liu2024probing}, and indicate that this probing system works better with larger models.

\subsection{Practical Usage of Closed-Data Contamination Detection Methods}

In the previous subsections, we reviewed the different types contamination detection methods. We now wish to give more concrete intuition in how these methods can be applied in practice.

Contamination detection techniques strongly differ in terms of model access requirement.\Cref{tab:model_access} summarizes closed-data detection techniques organized by the minimum \emph{level of access to the LLM} which is required. We define three levels of model access: \emph{black-box}, which merely needs prompting access ; \emph{gray-box}, which needs access to output tokens distributions ; and \emph{white-box} which needs to see model weights. All methods based on performance analysis only require a black-box access, making them convenient to use. Membership-inference attacks are equally split between black-box and gray-box access requirements. Model memorization techniques mainly rely on prompting and thus only need black-box access. Lastly, for model confidence methods, which often analyze output distributions, a gray-box or even white-box model access is needed.

\input{Tables/closed_data_training_stage}

We also highlight that closed-data contamination detection techniques may naturally be applied to different model training stages. In \Cref{tab:training_stage}, we indicate the most suitable training stage(s) where each method can be applied, among (i) pre-training, (ii) instruction-tuning and (iii) reinforcement learning from human feedback (RLHF). Prompt tuning is naturally tailored for RLHF models which can be prompted with any user query. Cloze tasks and likelihood-based methods are suitable for pre-trained models, which have captured tokens distributions. We emphasize that this categorization is flexible and not strictly defined: for instance, methods most suited for pre-trained models, like Min-K\% Prob, can in theory be applied to instruction-tuned or RLHF models. However, results should be cautiously interpreted as RLHF often smoothens overconfident predictions to align with human preferences, weakening the contamination signal.

%% file: Tables/closed_data_model_access.tex
\begin{table}[t]
\resizebox{0.99\columnwidth}{!}{
\begin{tabular}{llll}

\toprule 

\textbf{Closed-Data Method Type}
& \textbf{Black-box}   
& \textbf{Gray-box} 
& \textbf{White-box} \\

\midrule

\multirow{5}{*}{Performance Analysis} & CAP \cite{zhao2024cap} \\ 
& \citet{aiyappa-etal-2023-trust} \\ 
& \citet{roberts2023data} \\
& \citet{li2023task} \\
& \citet{cao2024concerned} \\

\midrule

\multirow{8}{*}{Membership Inference Attacks} & & RECALL \citep{xie2024recall} \\
& MaxNorm\citep{meeus2024did} \\ 
& & PREMIA \citep{feng2024exposing} \\ 
& MIA-Tuner \citep{fu2024mia} \\ 
& SPV-MIA \citep{fu2023practical} \\ 
& & CAMIA \citep{chang2024context} \\ 
& S2-MIA \citep{li2024generating} \\ 
& & EM-MIA \citep{kim2024detecting} \\ 

\midrule 

\multirow{11}{*}{Model Memorization} & & & \citet{magar2022data} \\ 
& \citet{carlini2023quantifying} \\ 
& \citet{Gemma2024OpenModels} \\
& & Dynamic Soft Prompting \citet{wang2024unlock-memorizing} \\ 
& \citet{nasr2023scalable} \\ 
& \emph{According To} prompting \citep{weller2023according} \\ 
& DCQ \citep{samuel2024towards} \\ 
& Local order quiz \citep{samuel2024towards} \\ 
& \citet{chang-etal-2023-speak} \\ 
& TS-Guessing \citep{deng2023investigating} \\ 
& \citet{ranaldi2024investigating} \\

\midrule 

\multirow{10}{*}{Model Confidence} & & Min-K\% Prob \citep{shi2024detecting} \\ 
& & DC-PDD \citep{zhang2024pretraining} \\ 
& & Min-K\%++ \citep{zhang2024min} \\ 
& & \citet{li2023estimating} \\ 
& & CDD and TED \citep{dong2024generalization} \\ 
& & \citet{xu2024benchmarking} \\ 
& & SURP \citep{zhang2024adaptive} \\ 
& \citet{oren2023proving} \\ 
& & & DICE \citep{tu2024dice} \\ 
& & & \citet{liu2024probing} \\ 

\bottomrule
\end{tabular}
}
\caption{\small Classification of closed-data contamination detection methods as per the level of access to the LLM required. Black-box methods simply need to prompt the model, while gray-box methods require access to the output tokens distributions and white-box techniques need full access to model weights.}
\label{tab:model_access}
\end{table}

%% file: Tables/closed_data_training_stage.tex
\begin{table}[t]
\resizebox{0.99\columnwidth}{!}{
\begin{tabular}{lllll}

\toprule 

\textbf{Closed-Data Method Type}
& \textbf{Contamination Detection Method}   
& \textbf{Pre-training}   
& \textbf{Instruction-tuning} 
& \textbf{RLHF} \\

\midrule

\multirow{5}{*}{Performance Analysis} 
& CAP \cite{zhao2024cap} & \cmark & \cmark & \\ 
& \citet{aiyappa-etal-2023-trust} & & \cmark & \cmark \\ 
& \citet{roberts2023data} & \cmark & & \\
& \citet{li2023task} & \cmark & & \\
& \citet{cao2024concerned} & \cmark & & \\

\midrule

\multirow{8}{*}{Membership Inference Attacks} 
& RECALL \citep{xie2024recall} & \cmark & & \\
& MaxNorm\citep{meeus2024did} & \cmark & & \\
& PREMIA \citep{feng2024exposing} & & & \cmark \\ 
& MIA-Tuner \citep{fu2024mia} & & \cmark & \\ 
& SPV-MIA \citep{fu2023practical} & & \cmark & \\ 
& CAMIA \citep{chang2024context} & \cmark & & \\
& EM-MIA \citep{kim2024detecting} & \cmark & & \\
& S2-MIA \citep{li2024generating} & & \cmark & \cmark \\ 

\midrule 

\multirow{11}{*}{Model Memorization} 
& \citet{magar2022data} & \cmark & & \\
& \citet{carlini2023quantifying} & \cmark & & \\
& \citet{Gemma2024OpenModels} & \cmark & & \\
& Dynamic Soft Prompting \citet{wang2024unlock-memorizing} & \cmark & & \\
& \citet{nasr2023scalable} & & & \cmark \\ 
& \emph{According To} prompting \citep{weller2023according} & & & \cmark \\ 
& DCQ \citep{samuel2024towards} & & & \cmark \\ 
& Local order quiz \citep{samuel2024towards} & & & \cmark \\ 
& \citet{chang-etal-2023-speak} & \cmark & & \\
& TS-Guessing \citep{deng2023investigating} & \cmark & & \\
& \citet{ranaldi2024investigating} & \cmark & & \\

\midrule 

\multirow{10}{*}{Model Confidence} 
& Min-K\% Prob \citep{shi2024detecting} & \cmark & & \\
& DC-PDD \citep{zhang2024pretraining} & \cmark & & \\
& Min-K\%++ \citep{zhang2024min} & \cmark & & \\
& \citet{li2023estimating} & \cmark & & \\
& CDD and TED \citep{dong2024generalization} & \cmark & & \\
& \citet{xu2024benchmarking} & \cmark & & \\
& SURP \citep{zhang2024adaptive} & \cmark & & \\
& \citet{oren2023proving} & \cmark & & \\
& DICE \citep{tu2024dice} & & \cmark & \\ 
& \citet{liu2024probing} & & \cmark & \\ 

\bottomrule
\end{tabular}
}
\caption{\small Classification of closed-data contamination detection techniques according to the model training stage(s) which they are more suitable to be applied to.}
\label{tab:training_stage}
\end{table}

%% file: Sections/5_discussion.tex
\section{Discussion}
\label{sec:discussion}

\subsection{Best Practices to Avoid Contamination}

Beyond detecting contaminated evaluation data points, recent works have called the community to adopt better practices to reduce contamination. 

\paragraph{Scanning newly released evaluation datasets}
\citet{sainz2023nlp} argue to develop automatic or semi-automatic approaches to detect contamination for new benchmarks and design mechanisms to flag those works with contamination. This process goes hand in hand with creating new, non-contaminated evaluation datasets.

\paragraph{Evaluating on a wide spectrum}
To comprehensively and fairly evaluate LLMs, \citet{zhou2023don} suggest to use more benchmarks from diverse sources, covering both basic and advanced capabilities. Furthermore, they recommend employing multiple task prompts for benchmark tests to derive model performance that is more stable and reliable. Additionally, they highlight the necessity of conducting data decontamination checks between pre-training data and any evaluation data when using benchmarks.

\paragraph{Encrypting evaluation datasets} 
\citet{jacovi2023stop} advocate for the encryption of publicly released test data using a public key, coupled with licensing agreements that prohibit derivative distribution. They recommend implementing training exclusion controls for closed APIs and safeguarding test data by refusing to perform evaluations without such controls. Moreover, they suggest avoiding data that appears with the solution and releasing internet-derived data along with the corresponding web-page context.

\paragraph{Not leaking data to closed-source APIs}
Aside from technical methodologies, \citet{balloccu2024leak} conduct a systematic literature review of 225 papers and carefully detail data leakage from them to closed-source models like the GPT-series. Overall, they conclude that $\sim$42\% of reviewed papers leaked data to GPT-3.5 and GPT-4 for a total of $\sim$4.7M benchmark samples across 263 benchmarks. The authors report evaluation malpractices and propose a list of suggested practices for evaluating closed-source LLMs. \newline 

Several data contamination mitigation techniques are detailed in the survey work of \citet{xu2024benchmark}.

\subsection{Towards Open-Source Pre-Training Data}

As we have seen in this study, access to the pre-training data is the first fundamental characterization of contamination detection technique. When the pre-training data are publicly available, the task of defining membership to the pre-training set becomes drastically simpler. To facilitate contamination detection, it is important for the research community to move towards the general usage of open-source pre-training sets.

\citet{duan2024membership} introduce the \textbf{MIMIR} dataset, constructed from The Pile training set \citep{gao2020pile}, on the which popular open-source models such as the Pythia \citep{biderman2023pythia} and GPT-Neo\footnote{\url{https://github.com/EleutherAI/gpt-neo}} model series are pre-trained. MIMIR covers domain-specific subsets (e.g. GitHub, Wikipedia, Arxiv) and extracts members through several n-gram sizes, allowing researchers to work on multiple levels of membership granularity. MIMIR has become widely used in MIA research with LLMs. 
\textbf{OLMoMIA} \citep{kim2024detecting} introduces a membership inference dataset centered around the open-source OLMo-7B LLM \citep{groeneveld2024olmo}. Knowing that a pre-training epoch roughly consists of 450k, training data seen before 100k pre-training steps is considered as member, and training data 400k and 450k steps is non-member.
\textbf{Dolma-Book} \citep{zhang2024adaptive} also leverages the publicly available pre-training data from OLMo, and samples non-member books from books from the Project Gutenberg\footnote{\url{https://www.gutenberg.org/}} dated after January 1st, 2024.

\subsection{New Evaluation Benchmarks}

Researchers have crafted new datasets specifically introduced with the aim to provide contamination-free LLM evaluation.

\paragraph{Benchmark Perturbation}
Several new datasets are crafted by \emph{perturbing} existing benchmarks, for instance through rephrasing instructions or adding more complex instructions. Perturbed data points are more challenging for LLMs, and can be designed to be adversarial examples aiming to fool the LLMs. Moreover, \citet{alzahrani2024benchmarks} point out that slight perturbations to existing benchmarks may disrupt model rankings on leaderboards, calling for better evaluation benchmarks. Such benchmarks include:

\begin{itemize}[leftmargin=*,itemsep=0.1em]
    \item \textbf{GSM-Plus} \citep{li2024gsm}. Using the 1,319 questions of the grade-school level maths word problems GSM8K dataset \citep{gsm8k} as seed dataset, the authors apply eight perturbations of five different types: numerical variation, arithmetic variation, problem understanding, distractor insertion and critical thinking. The dataset is generated with GPT-4 (manually verified by humans) and has 10,552 questions in total. High relative performance drop rates of above 30\% are observed for open-source models. Several elaborate prompting techniques such as chain-of-thought fall short of providing meaningful improvement, but a custom compositional prompting technique provides notable performance improvement.

    \item \textbf{MATH-Perturb} \citep{huang2023c} introduces two flavors of variations to 279 problems from the hardest level (level 5) of MATH dataset \citep{math}: slightly different problems solvable with the same method (MATH-P-Simple), and problems needing a different approach (MATH-P-HARD). Performance degrades notably for all LLMS on MATH-P-HARD. Interestingly, a lot of errors can be traced to memorization: the model simply blindly applies the same technique used to solve the original problem.
\end{itemize}

\paragraph{Dynamic Evaluation}
Some new datasets discard the \emph{static} paradigm of existing LLM evaluation datasets, and instead propose a \emph{dynamic}, ever-changing or periodically updated benchmark to avoid contamination. 

\begin{itemize}[leftmargin=*,itemsep=0.1em]
    \item \textbf{LatestEval} \citep{li2023avoiding} leverages the most recent texts to create dynamic reading comprehension evaluations. The benchmark is constructed in three steps: \Ni collect the latest texts; \Nii extract key information, and \Niii construct questions based on the extracted information through template-filling or LLMs.
    
    \item \textbf{KIEval} \citep{yu2024kieval} is an interactive evaluation framework incorporating an LLM-powered "interactor", which can ask follow-up questions in multi-round dialogue that leads to contamination-resilient evaluation. 
    
    \item \textbf{LiveCodeBench} \citep{jain2024livecodebench} continuously collects new problems from LeetCode, AtCoder and CodeForces to provide a contamination-free code generation benchmark. 

    \item \textbf{LiveBench} \citep{white2024livebench} is a challenging benchmark made of 18 tasks from 6 categories: math, coding, reasoning, language comprehension, instruction following, and data analysis. Tasks are more complicated versions of existing benchmarks, such as BigBench-Hard \citep{bigbench}, and are evaluated against an objective ground-truth rather than with human or LLM preference. To avoid contamination, one sixth of questions are replaced every month, chosen among oldest and easiest tasks.

    \item \textbf{Chatbot Arena} \citep{zheng2023judging} is a popular crowdsourced LLM evaluation platform. Humans submit queries and are shown answers from two LLMs which quality they have to rank. By construction, this benchmark is not set in stone and changes LLM rankings as more users interact with the platform. Besides, the design of Chatbot Arena limits contamination as user-generated prompts and model response rankings are less likely to be trained on. 
\end{itemize}

Several dynamic benchmarks are specifically crafted for membership-inference attacks:

\begin{itemize}[leftmargin=*,itemsep=0.1em]

    \item \textbf{WikiMIA} \citep{shi2024detecting} is a dynamic benchmark made of Wikipedia events created after model training (\ie\ after 2023). ChatGPT is leveraged to  paraphrase examples for evaluation, as also performed in CleanEval \citep{zhu2023clean}. 

    \item \textbf{WikiMIA-24} \citet{fu2024mia} tackle WikiMIA's as well as any publicly released dataset issue: cutoff dates will need to be adjusted as time goes on. To address the issue, the authors extend WikiMIA's cutoff to March 2024.

    \item \textbf{ArxivMIA} \citep{liu2024probing} gathers paper abstracts from the Computer Science and Maths sections of Arxiv. Abstracts dated after 2024 are considered non-members of LLMs pre-training sets.

    \item \textbf{PatentMIA} \citep{zhang2024pretraining} constructs a benchmarks where non-members are 5,000 patents from Google-Patents which are in Chinese language and dated after March 1st, 2024, and members are 5,000 such patents dated from before January 1st, 2023.
\end{itemize}

\paragraph{Protected Evaluation}
Another strategy to avoid contamination of evaluation data is to protect it by locking it out of the public domain. \textbf{Termite} \citep{ranaldi2024investigating} is a new text-to-SQL dataset, locked out of public access through search engines via an encryption key. Termite is made of handcrafted databases, each paired with around five queries, designed to match properties of the Spider dataset \citep{yu2018spider}, which shows high signs of contamination on GPT-3.5.

Scaling datasets used for contamination detection is critical, and such datasets should follow advancements in scale and complexity of datasets used to evaluate LLMs. \citet{samuel2024towards} emphasize that as of late 2024, there remains a wide discrepancy between datasets used to evaluate LLM performance and datasets used to detect contamination, with the latter usually lagging the former by several years.

\subsection{Future Challenges}

Given the fast and ever-changing landscape of machine learning research, future directions for contamination detection in LLMs could encompass a broad variety of methodologies and technological aspects. We highlight a few potentially critical areas of focus:

\paragraph{Real-time contamination detection systems} 
Real-time data contamination detection systems that can monitor data streams continuously and alert users to potential contamination events as they happen would be of upmost importance in critical applications like finance \citep{fresard2011pernicious} and healthcare \citep{gao2022robust} where data integrity is paramount and model reliability critical. Given the sheer volume of new data uploaded daily on the Internet, solving this task requires major technological breakthroughs. 
The key difficulty lies in achieving high recall, to ensure that no contaminated data sample slips through undetected.
We hypothesize that compression \citep{han2015deep}, distillation \citep{hinton2015distilling,sanh2019distilbert} and quantization \citep{banner2018scalable,sun2020ultra} techniques could play a key role in solving this problem, by decreasing model size, accelerating inference and scaling edge-device deployment of anomaly detection models or closed-data contamination detection techniques. 
Besides, leveraging federated learning and edge computing could enable decentralized, privacy-preserving detection at the source of data generation.

\paragraph{Bypassing contamination detection}
\citet{dekoninck2024evading} show a very effective way to bypass several existing contamination detection methods: Evasive Augmentation Learning (EAL) consists in paraphrasing benchmarks with GPT-4 and fine-tuning the LLM on the paraphrased data. This method can bypass even oracle-access contamination detection methods, calling for the development of more robust methods.
An intuitive future line of defense to such techniques is to incorporate large-scale adversarial training \citep{jia2017adversarial,liu2020adversarial} into contamination detection models, and train these models on benchmarks which have been paraphrased or noised. 
We expect to witness a \emph{cat-and-mouse} game between techniques such as EAL, which contaminate under the radar, and contamination detection techniques catching up to the finest transformations of benchmarks, such as paraphrases.

\paragraph{Ethical and legal data frameworks} 
Ethical and legal frameworks \citep{chang2021ethical} are needed to govern the collection, usage, and management of LLM pre-training data. 
A strong legal framework will help prevent the inclusion of data from unethical sources or from copyrighted sources, as well as help prevent the contamination of evaluation benchmarks into widely used pre-training data sources (e.g., CommonCrawl).
However, adopting such a worldwide pre-training data legal framework would be challenging due to several reasons.
First, the lines are blurry between public and private data, and many web-crawled datasets include a mix of public, open-access and proprietary data.
Jurisdiction surrounding data varies from a country to another, complicating the adoption of global standards.
Besides, the authorization to train machine learning models on copyrighted data remains unclear in many jurisdictions to this day. 
Another major blocker is the fact that pre-training data is nowadays treated with secrecy by large technological companies releasing flagship LLMs, as publishing the exact pre-training data would enable their competitors to catch up with their own LLMs.
Lastly, enforcing and auditing a legal framework surrounding pre-training pipelines in practice would be very complex and costly. 
Above all, legal frameworks on pre-training data are needed to protect personal data privacy. The General Data Protection Regulation (GDPR), acted in 2018 in the European Union (EU), imposes strict rules on how organizations can collect, store and use personal data from EU citizens. With the exponential growth of pre-training corpora, it becomes harder and harder yet critical to ensure that personal data does not get included into LLM training sets. Once an LLM has seen and memorized personal data, there is no clear mechanism for an individual to exercise their right to erase their personal data from the LLM's knowledge. In the near future, developing contamination detection techniques that do not compromise individual privacy will be crucial \citep{lin2010mdpa, hayes2018contamination}.

%% file: Sections/6_conclusion.tex
\section{Conclusion}

As LLMs rapidly evolve and the training data become more extensive, LLMs' performances are inevitably biased by contamination. 
Eventually, most (if not all) target-answer-based public evaluation datasets end up in commonly used pre-training data dumps, which are updated every few months.
Therefore, to help detect and quantify contamination, in this survey, we investigated, structured, and classified the current landscape of contamination detection in LLMs. 
Depending on a researcher's level of access to the data and to the LLM parameters, our survey can quickly guide them toward a set of relevant tools to detect contamination.

%% file: main.bbl
\begin{thebibliography}{153}
\providecommand{\natexlab}[1]{#1}
\providecommand{\url}[1]{\texttt{#1}}
\expandafter\ifx\csname urlstyle\endcsname\relax
  \providecommand{\doi}[1]{doi: #1}\else
  \providecommand{\doi}{doi: \begingroup \urlstyle{rm}\Url}\fi

\bibitem[Abaskohi et~al.(2023)Abaskohi, Rothe, and Yaghoobzadeh]{abaskohi2023lm}
Amirhossein Abaskohi, Sascha Rothe, and Yadollah Yaghoobzadeh.
\newblock Lm-cppf: Paraphrasing-guided data augmentation for contrastive prompt-based few-shot fine-tuning.
\newblock \emph{arXiv preprint arXiv:2305.18169}, 2023.

\bibitem[Achiam et~al.(2023)Achiam, Adler, Agarwal, Ahmad, Akkaya, Aleman, Almeida, Altenschmidt, Altman, Anadkat, et~al.]{achiam2023gpt}
Josh Achiam, Steven Adler, Sandhini Agarwal, Lama Ahmad, Ilge Akkaya, Florencia~Leoni Aleman, Diogo Almeida, Janko Altenschmidt, Sam Altman, Shyamal Anadkat, et~al.
\newblock Gpt-4 technical report.
\newblock \emph{arXiv preprint arXiv:2303.08774}, 2023.

\bibitem[Aiyappa et~al.(2023)Aiyappa, An, Kwak, and Ahn]{aiyappa-etal-2023-trust}
Rachith Aiyappa, Jisun An, Haewoon Kwak, and Yong-yeol Ahn.
\newblock Can we trust the evaluation on {C}hat{GPT}?
\newblock In Anaelia Ovalle, Kai-Wei Chang, Ninareh Mehrabi, Yada Pruksachatkun, Aram Galystan, Jwala Dhamala, Apurv Verma, Trista Cao, Anoop Kumar, and Rahul Gupta (eds.), \emph{Proceedings of the 3rd Workshop on Trustworthy Natural Language Processing (TrustNLP 2023)}, pp.\  47--54, Toronto, Canada, July 2023. Association for Computational Linguistics.
\newblock \doi{10.18653/v1/2023.trustnlp-1.5}.
\newblock URL \url{https://aclanthology.org/2023.trustnlp-1.5/}.

\bibitem[Alzahrani et~al.(2024)Alzahrani, Alyahya, Alnumay, Alrashed, Alsubaie, Almushaykeh, Mirza, Alotaibi, Altwairesh, Alowisheq, et~al.]{alzahrani2024benchmarks}
Norah Alzahrani, Hisham~Abdullah Alyahya, Yazeed Alnumay, Sultan Alrashed, Shaykhah Alsubaie, Yusef Almushaykeh, Faisal Mirza, Nouf Alotaibi, Nora Altwairesh, Areeb Alowisheq, et~al.
\newblock When benchmarks are targets: Revealing the sensitivity of large language model leaderboards.
\newblock \emph{arXiv preprint arXiv:2402.01781}, 2024.

\bibitem[Anil et~al.(2023)Anil, Dai, Firat, Johnson, Lepikhin, Passos, Shakeri, Taropa, Bailey, Chen, et~al.]{anil2023palm}
Rohan Anil, Andrew~M Dai, Orhan Firat, Melvin Johnson, Dmitry Lepikhin, Alexandre Passos, Siamak Shakeri, Emanuel Taropa, Paige Bailey, Zhifeng Chen, et~al.
\newblock Palm 2 technical report.
\newblock \emph{arXiv preprint arXiv:2305.10403}, 2023.

\bibitem[Austin et~al.(2021)Austin, Odena, Nye, Bosma, Michalewski, Dohan, Jiang, Cai, Terry, Le, et~al.]{mbpp}
Jacob Austin, Augustus Odena, Maxwell Nye, Maarten Bosma, Henryk Michalewski, David Dohan, Ellen Jiang, Carrie Cai, Michael Terry, Quoc Le, et~al.
\newblock Program synthesis with large language models.
\newblock \emph{arXiv preprint arXiv:2108.07732}, 2021.

\bibitem[Bai et~al.(2023)Bai, Bai, Chu, Cui, Dang, Deng, Fan, Ge, Han, Huang, et~al.]{bai2023qwen}
Jinze Bai, Shuai Bai, Yunfei Chu, Zeyu Cui, Kai Dang, Xiaodong Deng, Yang Fan, Wenbin Ge, Yu~Han, Fei Huang, et~al.
\newblock Qwen technical report.
\newblock \emph{arXiv preprint arXiv:2309.16609}, 2023.

\bibitem[Balloccu et~al.(2024)Balloccu, Schmidtov{\'a}, Lango, and Du{\v{s}}ek]{balloccu2024leak}
Simone Balloccu, Patr{\'\i}cia Schmidtov{\'a}, Mateusz Lango, and Ond{\v{r}}ej Du{\v{s}}ek.
\newblock Leak, cheat, repeat: Data contamination and evaluation malpractices in closed-source llms.
\newblock \emph{arXiv preprint arXiv:2402.03927}, 2024.

\bibitem[Banerjee \& Lavie(2005)Banerjee and Lavie]{banerjee-lavie-2005-meteor}
Satanjeev Banerjee and Alon Lavie.
\newblock {METEOR}: An automatic metric for {MT} evaluation with improved correlation with human judgments.
\newblock In Jade Goldstein, Alon Lavie, Chin-Yew Lin, and Clare Voss (eds.), \emph{Proceedings of the {ACL} Workshop on Intrinsic and Extrinsic Evaluation Measures for Machine Translation and/or Summarization}, pp.\  65--72, Ann Arbor, Michigan, June 2005. Association for Computational Linguistics.
\newblock URL \url{https://aclanthology.org/W05-0909}.

\bibitem[Banner et~al.(2018)Banner, Hubara, Hoffer, and Soudry]{banner2018scalable}
Ron Banner, Itay Hubara, Elad Hoffer, and Daniel Soudry.
\newblock Scalable methods for 8-bit training of neural networks.
\newblock \emph{Advances in neural information processing systems}, 31, 2018.

\bibitem[Betker et~al.(2023)Betker, Goh, Jing, Brooks, Wang, Li, Ouyang, Zhuang, Lee, Guo, et~al.]{betker2023improving}
James Betker, Gabriel Goh, Li~Jing, Tim Brooks, Jianfeng Wang, Linjie Li, Long Ouyang, Juntang Zhuang, Joyce Lee, Yufei Guo, et~al.
\newblock Improving image generation with better captions.
\newblock \emph{Computer Science. https://cdn. openai. com/papers/dall-e-3. pdf}, 2\penalty0 (3):\penalty0 8, 2023.

\bibitem[Bi et~al.(2024)Bi, Chen, Chen, Chen, Dai, Deng, Ding, Dong, Du, Fu, et~al.]{bi2024deepseek}
Xiao Bi, Deli Chen, Guanting Chen, Shanhuang Chen, Damai Dai, Chengqi Deng, Honghui Ding, Kai Dong, Qiushi Du, Zhe Fu, et~al.
\newblock Deepseek llm: Scaling open-source language models with longtermism.
\newblock \emph{arXiv preprint arXiv:2401.02954}, 2024.

\bibitem[Biderman et~al.(2023)Biderman, Schoelkopf, Anthony, Bradley, O’Brien, Hallahan, Khan, Purohit, Prashanth, Raff, et~al.]{biderman2023pythia}
Stella Biderman, Hailey Schoelkopf, Quentin~Gregory Anthony, Herbie Bradley, Kyle O’Brien, Eric Hallahan, Mohammad~Aflah Khan, Shivanshu Purohit, USVSN~Sai Prashanth, Edward Raff, et~al.
\newblock Pythia: A suite for analyzing large language models across training and scaling.
\newblock In \emph{International Conference on Machine Learning}, pp.\  2397--2430. PMLR, 2023.

\bibitem[Bisk et~al.(2020)Bisk, Zellers, Gao, Choi, et~al.]{bisk2020piqa}
Yonatan Bisk, Rowan Zellers, Jianfeng Gao, Yejin Choi, et~al.
\newblock Piqa: Reasoning about physical commonsense in natural language.
\newblock In \emph{Proceedings of the AAAI conference on artificial intelligence}, volume~34, pp.\  7432--7439, 2020.

\bibitem[Brown et~al.(2020)Brown, Mann, Ryder, Subbiah, Kaplan, Dhariwal, Neelakantan, Shyam, Sastry, Askell, et~al.]{brown2020language}
Tom Brown, Benjamin Mann, Nick Ryder, Melanie Subbiah, Jared~D Kaplan, Prafulla Dhariwal, Arvind Neelakantan, Pranav Shyam, Girish Sastry, Amanda Askell, et~al.
\newblock Language models are few-shot learners.
\newblock \emph{Advances in neural information processing systems}, 33:\penalty0 1877--1901, 2020.

\bibitem[Cai et~al.(2024)Cai, Cao, Chen, Chen, Chen, Chen, Chen, Chen, Chen, Chu, Dong, Duan, Fan, Fei, Gao, Ge, Gu, Gu, Gui, Guo, Guo, He, Hu, Huang, Jiang, Jiao, Jin, Lei, Li, Li, Li, Li, Li, Li, Liu, Liu, Hong, Liu, Liu, Liu, Lv, Lv, Lv, Ma, Ma, Ma, Ning, Ouyang, Qiu, Qu, Shang, Shao, Song, Song, Sui, Sun, Sun, Tang, Wang, Wang, Wang, Wang, Wang, Wang, Wang, Wei, Weng, Wu, Xiong, Xu, Xu, Yan, Yan, Yang, Ye, Ying, Yu, Yu, Zang, Zhang, Zhang, Zhang, Zhang, Zhang, Zhang, Zhang, Zhang, Zhang, Zhang, Zhang, Zhao, Zhao, Zhao, Zhou, Zhou, Zhuo, Zou, Qiu, Qiao, and Lin]{cai2024internlm2}
Zheng Cai, Maosong Cao, Haojiong Chen, Kai Chen, Keyu Chen, Xin Chen, Xun Chen, Zehui Chen, Zhi Chen, Pei Chu, Xiaoyi Dong, Haodong Duan, Qi~Fan, Zhaoye Fei, Yang Gao, Jiaye Ge, Chenya Gu, Yuzhe Gu, Tao Gui, Aijia Guo, Qipeng Guo, Conghui He, Yingfan Hu, Ting Huang, Tao Jiang, Penglong Jiao, Zhenjiang Jin, Zhikai Lei, Jiaxing Li, Jingwen Li, Linyang Li, Shuaibin Li, Wei Li, Yining Li, Hongwei Liu, Jiangning Liu, Jiawei Hong, Kaiwen Liu, Kuikun Liu, Xiaoran Liu, Chengqi Lv, Haijun Lv, Kai Lv, Li~Ma, Runyuan Ma, Zerun Ma, Wenchang Ning, Linke Ouyang, Jiantao Qiu, Yuan Qu, Fukai Shang, Yunfan Shao, Demin Song, Zifan Song, Zhihao Sui, Peng Sun, Yu~Sun, Huanze Tang, Bin Wang, Guoteng Wang, Jiaqi Wang, Jiayu Wang, Rui Wang, Yudong Wang, Ziyi Wang, Xingjian Wei, Qizhen Weng, Fan Wu, Yingtong Xiong, Chao Xu, Ruiliang Xu, Hang Yan, Yirong Yan, Xiaogui Yang, Haochen Ye, Huaiyuan Ying, Jia Yu, Jing Yu, Yuhang Zang, Chuyu Zhang, Li~Zhang, Pan Zhang, Peng Zhang, Ruijie Zhang, Shuo Zhang, Songyang Zhang, Wenjian Zhang,
  Wenwei Zhang, Xingcheng Zhang, Xinyue Zhang, Hui Zhao, Qian Zhao, Xiaomeng Zhao, Fengzhe Zhou, Zaida Zhou, Jingming Zhuo, Yicheng Zou, Xipeng Qiu, Yu~Qiao, and Dahua Lin.
\newblock Internlm2 technical report, 2024.

\bibitem[Cao et~al.(2024)Cao, Zhang, and Cheung]{cao2024concerned}
Jialun Cao, Wuqi Zhang, and Shing-Chi Cheung.
\newblock Concerned with data contamination? assessing countermeasures in code language model.
\newblock \emph{arXiv preprint arXiv:2403.16898}, 2024.

\bibitem[Carlini et~al.(2021)Carlini, Tramer, Wallace, Jagielski, Herbert-Voss, Lee, Roberts, Brown, Song, Erlingsson, et~al.]{carlini2021extracting}
Nicholas Carlini, Florian Tramer, Eric Wallace, Matthew Jagielski, Ariel Herbert-Voss, Katherine Lee, Adam Roberts, Tom Brown, Dawn Song, Ulfar Erlingsson, et~al.
\newblock Extracting training data from large language models.
\newblock In \emph{30th USENIX Security Symposium (USENIX Security 21)}, pp.\  2633--2650, 2021.

\bibitem[Carlini et~al.(2022)Carlini, Chien, Nasr, Song, Terzis, and Tramer]{carlini2022membership}
Nicholas Carlini, Steve Chien, Milad Nasr, Shuang Song, Andreas Terzis, and Florian Tramer.
\newblock Membership inference attacks from first principles.
\newblock In \emph{2022 IEEE Symposium on Security and Privacy (SP)}, pp.\  1897--1914. IEEE, 2022.

\bibitem[Carlini et~al.(2023)Carlini, Ippolito, Jagielski, Lee, Tramer, and Zhang]{carlini2023quantifying}
Nicholas Carlini, Daphne Ippolito, Matthew Jagielski, Katherine Lee, Florian Tramer, and Chiyuan Zhang.
\newblock Quantifying memorization across neural language models.
\newblock In \emph{The Eleventh International Conference on Learning Representations}, 2023.
\newblock URL \url{https://openreview.net/forum?id=TatRHT_1cK}.

\bibitem[Chang et~al.(2024)Chang, Shamsabadi, Katevas, Haddadi, and Shokri]{chang2024context}
Hongyan Chang, Ali~Shahin Shamsabadi, Kleomenis Katevas, Hamed Haddadi, and Reza Shokri.
\newblock Context-aware membership inference attacks against pre-trained large language models.
\newblock \emph{arXiv preprint arXiv:2409.13745}, 2024.

\bibitem[Chang et~al.(2023)Chang, Cramer, Soni, and Bamman]{chang-etal-2023-speak}
Kent Chang, Mackenzie Cramer, Sandeep Soni, and David Bamman.
\newblock Speak, memory: An archaeology of books known to {C}hat{GPT}/{GPT}-4.
\newblock In Houda Bouamor, Juan Pino, and Kalika Bali (eds.), \emph{Proceedings of the 2023 Conference on Empirical Methods in Natural Language Processing}, pp.\  7312--7327, Singapore, December 2023. Association for Computational Linguistics.
\newblock \doi{10.18653/v1/2023.emnlp-main.453}.
\newblock URL \url{https://aclanthology.org/2023.emnlp-main.453}.

\bibitem[Chang(2021)]{chang2021ethical}
Victor Chang.
\newblock An ethical framework for big data and smart cities.
\newblock \emph{Technological Forecasting and Social Change}, 165:\penalty0 120559, 2021.

\bibitem[Chen et~al.(2023)Chen, Jiao, Li, Qin, Ravaut, Zhao, Xiong, and Joty]{chen2023chatgpt}
Hailin Chen, Fangkai Jiao, Xingxuan Li, Chengwei Qin, Mathieu Ravaut, Ruochen Zhao, Caiming Xiong, and Shafiq Joty.
\newblock Chatgpt's one-year anniversary: Are open-source large language models catching up?
\newblock \emph{arXiv preprint arXiv:2311.16989}, 2023.

\bibitem[Chen et~al.(2021{\natexlab{a}})Chen, Tworek, Jun, Yuan, Pinto, Kaplan, Edwards, Burda, Joseph, Brockman, et~al.]{chen2021evaluating}
Mark Chen, Jerry Tworek, Heewoo Jun, Qiming Yuan, Henrique Ponde de~Oliveira Pinto, Jared Kaplan, Harri Edwards, Yuri Burda, Nicholas Joseph, Greg Brockman, et~al.
\newblock Evaluating large language models trained on code.
\newblock \emph{arXiv preprint arXiv:2107.03374}, 2021{\natexlab{a}}.

\bibitem[Chen et~al.(2021{\natexlab{b}})Chen, Chen, Smiley, Shah, Borova, Langdon, Moussa, Beane, Huang, Routledge, et~al.]{chen2021finqa}
Zhiyu Chen, Wenhu Chen, Charese Smiley, Sameena Shah, Iana Borova, Dylan Langdon, Reema Moussa, Matt Beane, Ting-Hao Huang, Bryan Routledge, et~al.
\newblock Finqa: A dataset of numerical reasoning over financial data.
\newblock \emph{arXiv preprint arXiv:2109.00122}, 2021{\natexlab{b}}.

\bibitem[Choi et~al.(2018)Choi, He, Iyyer, Yatskar, Yih, Choi, Liang, and Zettlemoyer]{quac}
Eunsol Choi, He~He, Mohit Iyyer, Mark Yatskar, Wen-tau Yih, Yejin Choi, Percy Liang, and Luke Zettlemoyer.
\newblock Quac: Question answering in context.
\newblock \emph{arXiv preprint arXiv:1808.07036}, 2018.

\bibitem[Chowdhery et~al.(2023)Chowdhery, Narang, Devlin, Bosma, Mishra, Roberts, Barham, Chung, Sutton, Gehrmann, et~al.]{chowdhery2023palm}
Aakanksha Chowdhery, Sharan Narang, Jacob Devlin, Maarten Bosma, Gaurav Mishra, Adam Roberts, Paul Barham, Hyung~Won Chung, Charles Sutton, Sebastian Gehrmann, et~al.
\newblock Palm: Scaling language modeling with pathways.
\newblock \emph{Journal of Machine Learning Research}, 24\penalty0 (240):\penalty0 1--113, 2023.

\bibitem[Clark et~al.(2019)Clark, Lee, Chang, Kwiatkowski, Collins, and Toutanova]{boolq}
Christopher Clark, Kenton Lee, Ming-Wei Chang, Tom Kwiatkowski, Michael Collins, and Kristina Toutanova.
\newblock Boolq: Exploring the surprising difficulty of natural yes/no questions.
\newblock \emph{arXiv preprint arXiv:1905.10044}, 2019.

\bibitem[Clark et~al.(2018)Clark, Cowhey, Etzioni, Khot, Sabharwal, Schoenick, and Tafjord]{arc}
Peter Clark, Isaac Cowhey, Oren Etzioni, Tushar Khot, Ashish Sabharwal, Carissa Schoenick, and Oyvind Tafjord.
\newblock Think you have solved question answering? try arc, the ai2 reasoning challenge.
\newblock \emph{arXiv preprint arXiv:1803.05457}, 2018.

\bibitem[Cobbe et~al.(2021)Cobbe, Kosaraju, Bavarian, Chen, Jun, Kaiser, Plappert, Tworek, Hilton, Nakano, et~al.]{gsm8k}
Karl Cobbe, Vineet Kosaraju, Mohammad Bavarian, Mark Chen, Heewoo Jun, Lukasz Kaiser, Matthias Plappert, Jerry Tworek, Jacob Hilton, Reiichiro Nakano, et~al.
\newblock Training verifiers to solve math word problems.
\newblock \emph{arXiv preprint arXiv:2110.14168}, 2021.

\bibitem[Dekoninck et~al.(2024)Dekoninck, M{\"u}ller, Baader, Fischer, and Vechev]{dekoninck2024evading}
Jasper Dekoninck, Mark~Niklas M{\"u}ller, Maximilian Baader, Marc Fischer, and Martin Vechev.
\newblock Evading data contamination detection for language models is (too) easy.
\newblock \emph{arXiv preprint arXiv:2402.02823}, 2024.

\bibitem[Deng et~al.(2023)Deng, Zhao, Tang, Gerstein, and Cohan]{deng2023investigating}
Chunyuan Deng, Yilun Zhao, Xiangru Tang, Mark Gerstein, and Arman Cohan.
\newblock Investigating data contamination in modern benchmarks for large language models.
\newblock \emph{arXiv preprint arXiv:2311.09783}, 2023.

\bibitem[Deng et~al.(2024)Deng, Zhao, Heng, Li, Cao, Tang, and Cohan]{deng2024unveiling}
Chunyuan Deng, Yilun Zhao, Yuzhao Heng, Yitong Li, Jiannan Cao, Xiangru Tang, and Arman Cohan.
\newblock Unveiling the spectrum of data contamination in language models: A survey from detection to remediation.
\newblock \emph{arXiv preprint arXiv:2406.14644}, 2024.

\bibitem[Devlin et~al.(2019)Devlin, Chang, Lee, and Toutanova]{devlin2019bert}
Jacob Devlin, Ming-Wei Chang, Kenton Lee, and Kristina Toutanova.
\newblock Bert: Pre-training of deep bidirectional transformers for language understanding.
\newblock In \emph{Proceedings of the 2019 conference of the North American chapter of the association for computational linguistics: human language technologies, volume 1 (long and short papers)}, pp.\  4171--4186, 2019.

\bibitem[Dodge et~al.(2021)Dodge, Sap, Marasovi{\'c}, Agnew, Ilharco, Groeneveld, Mitchell, and Gardner]{dodge2021documenting}
Jesse Dodge, Maarten Sap, Ana Marasovi{\'c}, William Agnew, Gabriel Ilharco, Dirk Groeneveld, Margaret Mitchell, and Matt Gardner.
\newblock Documenting large webtext corpora: A case study on the colossal clean crawled corpus.
\newblock \emph{arXiv preprint arXiv:2104.08758}, 2021.

\bibitem[Dong et~al.(2024)Dong, Jiang, Liu, Jin, and Li]{dong2024generalization}
Yihong Dong, Xue Jiang, Huanyu Liu, Zhi Jin, and Ge~Li.
\newblock Generalization or memorization: Data contamination and trustworthy evaluation for large language models.
\newblock \emph{arXiv preprint arXiv:2402.15938}, 2024.

\bibitem[Duan et~al.(2024)Duan, Suri, Mireshghallah, Min, Shi, Zettlemoyer, Tsvetkov, Choi, Evans, and Hajishirzi]{duan2024membership}
Michael Duan, Anshuman Suri, Niloofar Mireshghallah, Sewon Min, Weijia Shi, Luke Zettlemoyer, Yulia Tsvetkov, Yejin Choi, David Evans, and Hannaneh Hajishirzi.
\newblock Do membership inference attacks work on large language models?
\newblock \emph{arXiv preprint arXiv:2402.07841}, 2024.

\bibitem[Dubey et~al.(2024)Dubey, Jauhri, Pandey, Kadian, Al-Dahle, Letman, Mathur, Schelten, Yang, Fan, et~al.]{dubey2024llama}
Abhimanyu Dubey, Abhinav Jauhri, Abhinav Pandey, Abhishek Kadian, Ahmad Al-Dahle, Aiesha Letman, Akhil Mathur, Alan Schelten, Amy Yang, Angela Fan, et~al.
\newblock The llama 3 herd of models.
\newblock \emph{arXiv preprint arXiv:2407.21783}, 2024.

\bibitem[Elangovan et~al.(2021)Elangovan, He, and Verspoor]{elangovan2021memorization}
Aparna Elangovan, Jiayuan He, and Karin Verspoor.
\newblock Memorization vs. generalization: quantifying data leakage in nlp performance evaluation.
\newblock \emph{arXiv preprint arXiv:2102.01818}, 2021.

\bibitem[Feng et~al.(2024)Feng, Kasa, Yun, Teo, and Bodapati]{feng2024exposing}
Qizhang Feng, Siva~Rajesh Kasa, Hyokun Yun, Choon~Hui Teo, and Sravan~Babu Bodapati.
\newblock Exposing privacy gaps: Membership inference attack on preference data for llm alignment.
\newblock \emph{arXiv preprint arXiv:2407.06443}, 2024.

\bibitem[Fr{\'e}sard et~al.(2011)Fr{\'e}sard, P{\'e}rignon, and Wilhelmsson]{fresard2011pernicious}
Laurent Fr{\'e}sard, Christophe P{\'e}rignon, and Anders Wilhelmsson.
\newblock The pernicious effects of contaminated data in risk management.
\newblock \emph{Journal of Banking \& Finance}, 35\penalty0 (10):\penalty0 2569--2583, 2011.

\bibitem[Fu et~al.(2023)Fu, Wang, Gao, Liu, Li, and Jiang]{fu2023practical}
Wenjie Fu, Huandong Wang, Chen Gao, Guanghua Liu, Yong Li, and Tao Jiang.
\newblock Practical membership inference attacks against fine-tuned large language models via self-prompt calibration.
\newblock \emph{arXiv preprint arXiv:2311.06062}, 2023.

\bibitem[Fu et~al.(2024{\natexlab{a}})Fu, Wang, Gao, Liu, Li, and Jiang]{fu2024mia}
Wenjie Fu, Huandong Wang, Chen Gao, Guanghua Liu, Yong Li, and Tao Jiang.
\newblock Mia-tuner: Adapting large language models as pre-training text detector.
\newblock \emph{arXiv preprint arXiv:2408.08661}, 2024{\natexlab{a}}.

\bibitem[Fu et~al.(2024{\natexlab{b}})Fu, Uzuner, Yetisgen, and Xia]{fu2024does}
Yujuan Fu, Ozlem Uzuner, Meliha Yetisgen, and Fei Xia.
\newblock Does data contamination detection work (well) for llms? a survey and evaluation on detection assumptions.
\newblock \emph{arXiv preprint arXiv:2410.18966}, 2024{\natexlab{b}}.

\bibitem[Gao et~al.(2020)Gao, Biderman, Black, Golding, Hoppe, Foster, Phang, He, Thite, Nabeshima, et~al.]{gao2020pile}
Leo Gao, Stella Biderman, Sid Black, Laurence Golding, Travis Hoppe, Charles Foster, Jason Phang, Horace He, Anish Thite, Noa Nabeshima, et~al.
\newblock The pile: An 800gb dataset of diverse text for language modeling.
\newblock \emph{arXiv preprint arXiv:2101.00027}, 2020.

\bibitem[Gao et~al.(2022)Gao, Huang, Zhao, Cao, and Jiang]{gao2022robust}
Xuehong Gao, Guozhong Huang, Qiuhong Zhao, Cejun Cao, and Huiling Jiang.
\newblock Robust optimization model for medical staff rebalancing problem with data contamination during covid-19 pandemic.
\newblock \emph{International Journal of Production Research}, 60\penalty0 (5):\penalty0 1737--1766, 2022.

\bibitem[{Gemma Team Google DeepMind}(2024)]{Gemma2024OpenModels}
{Gemma Team Google DeepMind}.
\newblock {Gemma: Open Models Based on Gemini Research and Technology}.
\newblock Technical report, Google DeepMind, 2024.
\newblock Available online at gemma-1-report@google.com.

\bibitem[Golchin \& Surdeanu(2023{\natexlab{a}})Golchin and Surdeanu]{golchin2023data}
Shahriar Golchin and Mihai Surdeanu.
\newblock Data contamination quiz: A tool to detect and estimate contamination in large language models.
\newblock \emph{arXiv preprint arXiv:2311.06233}, 2023{\natexlab{a}}.

\bibitem[Golchin \& Surdeanu(2023{\natexlab{b}})Golchin and Surdeanu]{golchin2023time}
Shahriar Golchin and Mihai Surdeanu.
\newblock Time travel in llms: Tracing data contamination in large language models.
\newblock \emph{arXiv preprint arXiv:2308.08493}, 2023{\natexlab{b}}.

\bibitem[Groeneveld et~al.(2024)Groeneveld, Beltagy, Walsh, Bhagia, Kinney, Tafjord, Jha, Ivison, Magnusson, Wang, et~al.]{groeneveld2024olmo}
Dirk Groeneveld, Iz~Beltagy, Pete Walsh, Akshita Bhagia, Rodney Kinney, Oyvind Tafjord, Ananya~Harsh Jha, Hamish Ivison, Ian Magnusson, Yizhong Wang, et~al.
\newblock Olmo: Accelerating the science of language models.
\newblock \emph{arXiv preprint arXiv:2402.00838}, 2024.

\bibitem[Gunasekar et~al.(2023)Gunasekar, Zhang, Aneja, Mendes, Del~Giorno, Gopi, Javaheripi, Kauffmann, de~Rosa, Saarikivi, et~al.]{gunasekar2023textbooks}
Suriya Gunasekar, Yi~Zhang, Jyoti Aneja, Caio C{\'e}sar~Teodoro Mendes, Allie Del~Giorno, Sivakanth Gopi, Mojan Javaheripi, Piero Kauffmann, Gustavo de~Rosa, Olli Saarikivi, et~al.
\newblock Textbooks are all you need.
\newblock \emph{arXiv preprint arXiv:2306.11644}, 2023.

\bibitem[Han et~al.(2015)Han, Mao, and Dally]{han2015deep}
Song Han, Huizi Mao, and William~J Dally.
\newblock Deep compression: Compressing deep neural networks with pruning, trained quantization and huffman coding.
\newblock \emph{arXiv preprint arXiv:1510.00149}, 2015.

\bibitem[Hayes \& Ohrimenko(2018)Hayes and Ohrimenko]{hayes2018contamination}
Jamie Hayes and Olga Ohrimenko.
\newblock Contamination attacks and mitigation in multi-party machine learning.
\newblock \emph{Advances in neural information processing systems}, 31, 2018.

\bibitem[Hendrycks et~al.(2020)Hendrycks, Burns, Basart, Zou, Mazeika, Song, and Steinhardt]{mmlu}
Dan Hendrycks, Collin Burns, Steven Basart, Andy Zou, Mantas Mazeika, Dawn Song, and Jacob Steinhardt.
\newblock Measuring massive multitask language understanding.
\newblock \emph{arXiv preprint arXiv:2009.03300}, 2020.

\bibitem[Hendrycks et~al.(2021)Hendrycks, Burns, Kadavath, Arora, Basart, Tang, Song, and Steinhardt]{math}
Dan Hendrycks, Collin Burns, Saurav Kadavath, Akul Arora, Steven Basart, Eric Tang, Dawn Song, and Jacob Steinhardt.
\newblock Measuring mathematical problem solving with the math dataset.
\newblock \emph{arXiv preprint arXiv:2103.03874}, 2021.

\bibitem[Hermann et~al.(2015)Hermann, Kocisky, Grefenstette, Espeholt, Kay, Suleyman, and Blunsom]{cnndm}
Karl~Moritz Hermann, Tomas Kocisky, Edward Grefenstette, Lasse Espeholt, Will Kay, Mustafa Suleyman, and Phil Blunsom.
\newblock Teaching machines to read and comprehend.
\newblock \emph{Advances in neural information processing systems}, 28, 2015.

\bibitem[Hill et~al.(2015)Hill, Bordes, Chopra, and Weston]{hill2015goldilocks}
Felix Hill, Antoine Bordes, Sumit Chopra, and Jason Weston.
\newblock The goldilocks principle: Reading children's books with explicit memory representations.
\newblock \emph{arXiv preprint arXiv:1511.02301}, 2015.

\bibitem[Hinton et~al.(2015)Hinton, Vinyals, and Dean]{hinton2015distilling}
Geoffrey Hinton, Oriol Vinyals, and Jeff Dean.
\newblock Distilling the knowledge in a neural network.
\newblock \emph{arXiv preprint arXiv:1503.02531}, 2015.

\bibitem[Huang et~al.(2023)Huang, Bai, Zhu, Zhang, Zhang, Su, Liu, Lv, Zhang, Fu, et~al.]{huang2023c}
Yuzhen Huang, Yuzhuo Bai, Zhihao Zhu, Junlei Zhang, Jinghan Zhang, Tangjun Su, Junteng Liu, Chuancheng Lv, Yikai Zhang, Yao Fu, et~al.
\newblock C-eval: A multi-level multi-discipline chinese evaluation suite for foundation models.
\newblock \emph{Advances in Neural Information Processing Systems}, 36:\penalty0 62991--63010, 2023.

\bibitem[Hui et~al.(2024)Hui, Yang, Cui, Yang, Liu, Zhang, Liu, Zhang, Yu, Lu, et~al.]{hui2024qwen2}
Binyuan Hui, Jian Yang, Zeyu Cui, Jiaxi Yang, Dayiheng Liu, Lei Zhang, Tianyu Liu, Jiajun Zhang, Bowen Yu, Keming Lu, et~al.
\newblock Qwen2. 5-coder technical report.
\newblock \emph{arXiv preprint arXiv:2409.12186}, 2024.

\bibitem[Jacovi et~al.(2023)Jacovi, Caciularu, Goldman, and Goldberg]{jacovi2023stop}
Alon Jacovi, Avi Caciularu, Omer Goldman, and Yoav Goldberg.
\newblock Stop uploading test data in plain text: Practical strategies for mitigating data contamination by evaluation benchmarks.
\newblock \emph{arXiv preprint arXiv:2305.10160}, 2023.

\bibitem[Jain et~al.(2024)Jain, Han, Gu, Li, Yan, Zhang, Wang, Solar-Lezama, Sen, and Stoica]{jain2024livecodebench}
Naman Jain, King Han, Alex Gu, Wen-Ding Li, Fanjia Yan, Tianjun Zhang, Sida Wang, Armando Solar-Lezama, Koushik Sen, and Ion Stoica.
\newblock Livecodebench: Holistic and contamination free evaluation of large language models for code.
\newblock \emph{arXiv preprint arXiv:2403.07974}, 2024.

\bibitem[Jia \& Liang(2017)Jia and Liang]{jia2017adversarial}
Robin Jia and Percy Liang.
\newblock Adversarial examples for evaluating reading comprehension systems.
\newblock \emph{arXiv preprint arXiv:1707.07328}, 2017.

\bibitem[Jiang et~al.(2024{\natexlab{a}})Jiang, Sablayrolles, Roux, Mensch, Savary, Bamford, Chaplot, Casas, Hanna, Bressand, et~al.]{jiang2024mixtral}
Albert~Q Jiang, Alexandre Sablayrolles, Antoine Roux, Arthur Mensch, Blanche Savary, Chris Bamford, Devendra~Singh Chaplot, Diego de~las Casas, Emma~Bou Hanna, Florian Bressand, et~al.
\newblock Mixtral of experts.
\newblock \emph{arXiv preprint arXiv:2401.04088}, 2024{\natexlab{a}}.

\bibitem[Jiang et~al.(2024{\natexlab{b}})Jiang, Liu, Zhong, Schaeffer, Ouyang, Han, and Koyejo]{jiang2024investigating}
Minhao Jiang, Ken~Ziyu Liu, Ming Zhong, Rylan Schaeffer, Siru Ouyang, Jiawei Han, and Sanmi Koyejo.
\newblock Investigating data contamination for pre-training language models.
\newblock \emph{arXiv preprint arXiv:2401.06059}, 2024{\natexlab{b}}.

\bibitem[Joshi et~al.(2017)Joshi, Choi, Weld, and Zettlemoyer]{joshi2017triviaqa}
Mandar Joshi, Eunsol Choi, Daniel~S Weld, and Luke Zettlemoyer.
\newblock Triviaqa: A large scale distantly supervised challenge dataset for reading comprehension.
\newblock \emph{arXiv preprint arXiv:1705.03551}, 2017.

\bibitem[Kim et~al.(2024)Kim, Li, Spiliopoulou, Ma, Ballesteros, and Wang]{kim2024detecting}
Gyuwan Kim, Yang Li, Evangelia Spiliopoulou, Jie Ma, Miguel Ballesteros, and William~Yang Wang.
\newblock Detecting training data of large language models via expectation maximization.
\newblock \emph{arXiv preprint arXiv:2410.07582}, 2024.

\bibitem[Kocetkov et~al.(2022)Kocetkov, Li, Allal, Li, Mou, Ferrandis, Jernite, Mitchell, Hughes, Wolf, et~al.]{kocetkov2022stack}
Denis Kocetkov, Raymond Li, Loubna~Ben Allal, Jia Li, Chenghao Mou, Carlos~Mu{\~n}oz Ferrandis, Yacine Jernite, Margaret Mitchell, Sean Hughes, Thomas Wolf, et~al.
\newblock The stack: 3 tb of permissively licensed source code.
\newblock \emph{arXiv preprint arXiv:2211.15533}, 2022.

\bibitem[Kwiatkowski et~al.(2019)Kwiatkowski, Palomaki, Redfield, Collins, Parikh, Alberti, Epstein, Polosukhin, Devlin, Lee, et~al.]{kwiatkowski2019natural}
Tom Kwiatkowski, Jennimaria Palomaki, Olivia Redfield, Michael Collins, Ankur Parikh, Chris Alberti, Danielle Epstein, Illia Polosukhin, Jacob Devlin, Kenton Lee, et~al.
\newblock Natural questions: a benchmark for question answering research.
\newblock \emph{Transactions of the Association for Computational Linguistics}, 7:\penalty0 453--466, 2019.

\bibitem[Lee et~al.(2023)Lee, Hunter, and Ruiz]{lee2023platypus}
Ariel~N Lee, Cole~J Hunter, and Nataniel Ruiz.
\newblock Platypus: Quick, cheap, and powerful refinement of llms.
\newblock \emph{arXiv preprint arXiv:2308.07317}, 2023.

\bibitem[Lester et~al.(2021)Lester, Al-Rfou, and Constant]{lester2021power}
Brian Lester, Rami Al-Rfou, and Noah Constant.
\newblock The power of scale for parameter-efficient prompt tuning.
\newblock \emph{arXiv preprint arXiv:2104.08691}, 2021.

\bibitem[Levesque et~al.(2012)Levesque, Davis, and Morgenstern]{winograd}
Hector Levesque, Ernest Davis, and Leora Morgenstern.
\newblock The winograd schema challenge.
\newblock In \emph{Thirteenth international conference on the principles of knowledge representation and reasoning}, 2012.

\bibitem[Li \& Flanigan(2023)Li and Flanigan]{li2023task}
Changmao Li and Jeffrey Flanigan.
\newblock Task contamination: Language models may not be few-shot anymore.
\newblock \emph{arXiv preprint arXiv:2312.16337}, 2023.

\bibitem[Li et~al.(2024{\natexlab{a}})Li, Cui, Zhao, Kong, and Bi]{li2024gsm}
Qintong Li, Leyang Cui, Xueliang Zhao, Lingpeng Kong, and Wei Bi.
\newblock Gsm-plus: A comprehensive benchmark for evaluating the robustness of llms as mathematical problem solvers.
\newblock \emph{arXiv preprint arXiv:2402.19255}, 2024{\natexlab{a}}.

\bibitem[Li et~al.(2023{\natexlab{a}})Li, Zhao, Chia, Ding, Bing, Joty, and Poria]{li2023chain}
Xingxuan Li, Ruochen Zhao, Yew~Ken Chia, Bosheng Ding, Lidong Bing, Shafiq Joty, and Soujanya Poria.
\newblock Chain of knowledge: A framework for grounding large language models with structured knowledge bases.
\newblock \emph{arXiv preprint arXiv:2305.13269}, 2023{\natexlab{a}}.

\bibitem[Li(2023{\natexlab{a}})]{li2023estimating}
Yucheng Li.
\newblock Estimating contamination via perplexity: Quantifying memorisation in language model evaluation.
\newblock \emph{arXiv preprint arXiv:2309.10677}, 2023{\natexlab{a}}.

\bibitem[Li(2023{\natexlab{b}})]{li2023open}
Yucheng Li.
\newblock An open source data contamination report for llama series models.
\newblock \emph{arXiv preprint arXiv:2310.17589}, 2023{\natexlab{b}}.

\bibitem[Li et~al.(2023{\natexlab{b}})Li, Geurin, and Lin]{li2023avoiding}
Yucheng Li, Frank Geurin, and Chenghua Lin.
\newblock Avoiding data contamination in language model evaluation: Dynamic test construction with latest materials.
\newblock \emph{arXiv preprint arXiv:2312.12343}, 2023{\natexlab{b}}.

\bibitem[Li et~al.(2024{\natexlab{b}})Li, Liu, Wang, and Yang]{li2024generating}
Yuying Li, Gaoyang Liu, Chen Wang, and Yang Yang.
\newblock Generating is believing: Membership inference attacks against retrieval-augmented generation.
\newblock \emph{arXiv preprint arXiv:2406.19234}, 2024{\natexlab{b}}.

\bibitem[Lin et~al.(2021)Lin, Hilton, and Evans]{truthfulqa}
Stephanie Lin, Jacob Hilton, and Owain Evans.
\newblock Truthfulqa: Measuring how models mimic human falsehoods.
\newblock \emph{arXiv preprint arXiv:2109.07958}, 2021.

\bibitem[Lin et~al.(2010)Lin, Lu, and Shen]{lin2010mdpa}
Xiaodong Lin, Rongxing Lu, and Xuemin Shen.
\newblock Mdpa: multidimensional privacy-preserving aggregation scheme for wireless sensor networks.
\newblock \emph{Wireless Communications and Mobile Computing}, 10\penalty0 (6):\penalty0 843--856, 2010.

\bibitem[Liu et~al.(2020)Liu, Cheng, He, Chen, Wang, Poon, and Gao]{liu2020adversarial}
Xiaodong Liu, Hao Cheng, Pengcheng He, Weizhu Chen, Yu~Wang, Hoifung Poon, and Jianfeng Gao.
\newblock Adversarial training for large neural language models.
\newblock \emph{arXiv preprint arXiv:2004.08994}, 2020.

\bibitem[Liu et~al.(2024)Liu, Zhu, Tan, Lu, Liu, and Chen]{liu2024probing}
Zhenhua Liu, Tong Zhu, Chuanyuan Tan, Haonan Lu, Bing Liu, and Wenliang Chen.
\newblock Probing language models for pre-training data detection.
\newblock \emph{arXiv preprint arXiv:2406.01333}, 2024.

\bibitem[Maas et~al.(2011)Maas, Daly, Pham, Huang, Ng, and Potts]{maas2011learning}
Andrew Maas, Raymond~E Daly, Peter~T Pham, Dan Huang, Andrew~Y Ng, and Christopher Potts.
\newblock Learning word vectors for sentiment analysis.
\newblock In \emph{Proceedings of the 49th annual meeting of the association for computational linguistics: Human language technologies}, pp.\  142--150, 2011.

\bibitem[Maertens et~al.(2022)Maertens, Van~Petegem, Strijbol, Baeyens, Jacobs, Dawyndt, and Mesuere]{maertens2022dolos}
Rien Maertens, Charlotte Van~Petegem, Niko Strijbol, Toon Baeyens, Arne~Carla Jacobs, Peter Dawyndt, and Bart Mesuere.
\newblock Dolos: Language-agnostic plagiarism detection in source code.
\newblock \emph{Journal of Computer Assisted Learning}, 38\penalty0 (4):\penalty0 1046--1061, 2022.

\bibitem[Magar \& Schwartz(2022)Magar and Schwartz]{magar2022data}
Inbal Magar and Roy Schwartz.
\newblock Data contamination: From memorization to exploitation.
\newblock \emph{arXiv preprint arXiv:2203.08242}, 2022.

\bibitem[Maini et~al.(2025)Maini, Jia, Papernot, and Dziedzic]{maini2025llm}
Pratyush Maini, Hengrui Jia, Nicolas Papernot, and Adam Dziedzic.
\newblock Llm dataset inference: Did you train on my dataset?
\newblock \emph{Advances in Neural Information Processing Systems}, 37:\penalty0 124069--124092, 2025.

\bibitem[Meeus et~al.(2024{\natexlab{a}})Meeus, Jain, Rei, and de~Montjoye]{meeus2024did}
Matthieu Meeus, Shubham Jain, Marek Rei, and Yves-Alexandre de~Montjoye.
\newblock Did the neurons read your book? document-level membership inference for large language models.
\newblock In \emph{33rd USENIX Security Symposium (USENIX Security 24)}, pp.\  2369--2385, 2024{\natexlab{a}}.

\bibitem[Meeus et~al.(2024{\natexlab{b}})Meeus, Shilov, Jain, Faysse, Rei, and de~Montjoye]{meeus2024sok}
Matthieu Meeus, Igor Shilov, Shubham Jain, Manuel Faysse, Marek Rei, and Yves-Alexandre de~Montjoye.
\newblock Sok: Membership inference attacks on llms are rushing nowhere (and how to fix it).
\newblock \emph{arXiv preprint arXiv:2406.17975}, 2024{\natexlab{b}}.

\bibitem[Merity et~al.(2016)Merity, Xiong, Bradbury, and Socher]{merity2016pointer}
Stephen Merity, Caiming Xiong, James Bradbury, and Richard Socher.
\newblock Pointer sentinel mixture models.
\newblock \emph{arXiv preprint arXiv:1609.07843}, 2016.

\bibitem[Mihaylov et~al.(2018)Mihaylov, Clark, Khot, and Sabharwal]{openbookqa}
Todor Mihaylov, Peter Clark, Tushar Khot, and Ashish Sabharwal.
\newblock Can a suit of armor conduct electricity? a new dataset for open book question answering.
\newblock \emph{arXiv preprint arXiv:1809.02789}, 2018.

\bibitem[Narayan et~al.(2018)Narayan, Cohen, and Lapata]{xsum}
Shashi Narayan, Shay~B Cohen, and Mirella Lapata.
\newblock Don't give me the details, just the summary! topic-aware convolutional neural networks for extreme summarization.
\newblock \emph{arXiv preprint arXiv:1808.08745}, 2018.

\bibitem[Nasr et~al.(2023)Nasr, Carlini, Hayase, Jagielski, Cooper, Ippolito, Choquette-Choo, Wallace, Tram{\`e}r, and Lee]{nasr2023scalable}
Milad Nasr, Nicholas Carlini, Jonathan Hayase, Matthew Jagielski, A~Feder Cooper, Daphne Ippolito, Christopher~A Choquette-Choo, Eric Wallace, Florian Tram{\`e}r, and Katherine Lee.
\newblock Scalable extraction of training data from (production) language models.
\newblock \emph{arXiv preprint arXiv:2311.17035}, 2023.

\bibitem[Nijkamp et~al.(2022)Nijkamp, Pang, Hayashi, Tu, Wang, Zhou, Savarese, and Xiong]{nijkamp2022codegen}
Erik Nijkamp, Bo~Pang, Hiroaki Hayashi, Lifu Tu, Huan Wang, Yingbo Zhou, Silvio Savarese, and Caiming Xiong.
\newblock Codegen: An open large language model for code with multi-turn program synthesis.
\newblock \emph{arXiv preprint arXiv:2203.13474}, 2022.

\bibitem[Oren et~al.(2023)Oren, Meister, Chatterji, Ladhak, and Hashimoto]{oren2023proving}
Yonatan Oren, Nicole Meister, Niladri Chatterji, Faisal Ladhak, and Tatsunori~B Hashimoto.
\newblock Proving test set contamination in black box language models.
\newblock \emph{arXiv preprint arXiv:2310.17623}, 2023.

\bibitem[Ouyang et~al.(2022{\natexlab{a}})Ouyang, Wu, Jiang, Almeida, Wainwright, Mishkin, Zhang, Agarwal, Slama, Ray, et~al.]{instructgpt}
Long Ouyang, Jeffrey Wu, Xu~Jiang, Diogo Almeida, Carroll Wainwright, Pamela Mishkin, Chong Zhang, Sandhini Agarwal, Katarina Slama, Alex Ray, et~al.
\newblock Training language models to follow instructions with human feedback.
\newblock \emph{Advances in neural information processing systems}, 35:\penalty0 27730--27744, 2022{\natexlab{a}}.

\bibitem[Ouyang et~al.(2022{\natexlab{b}})Ouyang, Wu, Jiang, Almeida, Wainwright, Mishkin, Zhang, Agarwal, Slama, Ray, et~al.]{ouyang2022training}
Long Ouyang, Jeffrey Wu, Xu~Jiang, Diogo Almeida, Carroll Wainwright, Pamela Mishkin, Chong Zhang, Sandhini Agarwal, Katarina Slama, Alex Ray, et~al.
\newblock Training language models to follow instructions with human feedback.
\newblock \emph{Advances in neural information processing systems}, 35:\penalty0 27730--27744, 2022{\natexlab{b}}.

\bibitem[{\"{O}}zdayi et~al.(2023){\"{O}}zdayi, Peris, FitzGerald, Dupuy, Majmudar, Khan, Parikh, and Gupta]{mustafa2023memorized-prompt-tuning}
Mustafa {\"{O}}zdayi, Charith Peris, Jack FitzGerald, Christophe Dupuy, Jimit Majmudar, Haidar Khan, Rahil Parikh, and Rahul Gupta.
\newblock Controlling the extraction of memorized data from large language models via prompt-tuning.
\newblock In \emph{Proceedings of the 61st Annual Meeting of the Association for Computational Linguistics (Volume 2: Short Papers), {ACL} 2023, Toronto, Canada, July 9-14, 2023}, pp.\  1512--1521. Association for Computational Linguistics, 2023.

\bibitem[Palavalli et~al.(2024)Palavalli, Bertsch, and Gormley]{palavalli2024taxonomy}
Medha Palavalli, Amanda Bertsch, and Matthew~R Gormley.
\newblock A taxonomy for data contamination in large language models.
\newblock \emph{arXiv preprint arXiv:2407.08716}, 2024.

\bibitem[Paperno et~al.(2016)Paperno, Kruszewski, Lazaridou, Pham, Bernardi, Pezzelle, Baroni, Boleda, and Fern{\'a}ndez]{lambada}
Denis Paperno, Germ{\'a}n Kruszewski, Angeliki Lazaridou, Quan~Ngoc Pham, Raffaella Bernardi, Sandro Pezzelle, Marco Baroni, Gemma Boleda, and Raquel Fern{\'a}ndez.
\newblock The lambada dataset: Word prediction requiring a broad discourse context.
\newblock \emph{arXiv preprint arXiv:1606.06031}, 2016.

\bibitem[Radford et~al.(2019)Radford, Wu, Child, Luan, Amodei, Sutskever, et~al.]{radford2019language}
Alec Radford, Jeffrey Wu, Rewon Child, David Luan, Dario Amodei, Ilya Sutskever, et~al.
\newblock Language models are unsupervised multitask learners.
\newblock \emph{OpenAI blog}, 1\penalty0 (8):\penalty0 9, 2019.

\bibitem[Rafailov et~al.(2023)Rafailov, Sharma, Mitchell, Manning, Ermon, and Finn]{rafailov2023direct}
Rafael Rafailov, Archit Sharma, Eric Mitchell, Christopher~D Manning, Stefano Ermon, and Chelsea Finn.
\newblock Direct preference optimization: Your language model is secretly a reward model.
\newblock \emph{Advances in Neural Information Processing Systems}, 36:\penalty0 53728--53741, 2023.

\bibitem[Raffel et~al.(2020)Raffel, Shazeer, Roberts, Lee, Narang, Matena, Zhou, Li, and Liu]{commoncrawl}
Colin Raffel, Noam Shazeer, Adam Roberts, Katherine Lee, Sharan Narang, Michael Matena, Yanqi Zhou, Wei Li, and Peter~J. Liu.
\newblock Exploring the limits of transfer learning with a unified text-to-text transformer.
\newblock \emph{J. Mach. Learn. Res.}, 21\penalty0 (1), jan 2020.
\newblock ISSN 1532-4435.

\bibitem[Rajpurkar et~al.(2016)Rajpurkar, Zhang, Lopyrev, and Liang]{squad}
Pranav Rajpurkar, Jian Zhang, Konstantin Lopyrev, and Percy Liang.
\newblock Squad: 100,000+ questions for machine comprehension of text.
\newblock \emph{arXiv preprint arXiv:1606.05250}, 2016.

\bibitem[Rajpurkar et~al.(2018)Rajpurkar, Jia, and Liang]{squad2}
Pranav Rajpurkar, Robin Jia, and Percy Liang.
\newblock Know what you don't know: Unanswerable questions for squad.
\newblock \emph{arXiv preprint arXiv:1806.03822}, 2018.

\bibitem[Ranaldi et~al.(2024)Ranaldi, Ruzzetti, Onorati, Ranaldi, Giannone, Favalli, Romagnoli, and Zanzotto]{ranaldi2024investigating}
Federico Ranaldi, Elena~Sofia Ruzzetti, Dario Onorati, Leonardo Ranaldi, Cristina Giannone, Andrea Favalli, Raniero Romagnoli, and Fabio~Massimo Zanzotto.
\newblock Investigating the impact of data contamination of large language models in text-to-sql translation.
\newblock \emph{arXiv preprint arXiv:2402.08100}, 2024.

\bibitem[Reddy et~al.(2019)Reddy, Chen, and Manning]{coqa}
Siva Reddy, Danqi Chen, and Christopher~D Manning.
\newblock Coqa: A conversational question answering challenge.
\newblock \emph{Transactions of the Association for Computational Linguistics}, 7:\penalty0 249--266, 2019.

\bibitem[Reimers \& Gurevych(2019)Reimers and Gurevych]{reimers2019sentence}
Nils Reimers and Iryna Gurevych.
\newblock Sentence-bert: Sentence embeddings using siamese bert-networks.
\newblock \emph{arXiv preprint arXiv:1908.10084}, 2019.

\bibitem[Riddell et~al.(2024)Riddell, Ni, and Cohan]{riddell2024quantifying}
Martin Riddell, Ansong Ni, and Arman Cohan.
\newblock Quantifying contamination in evaluating code generation capabilities of language models.
\newblock \emph{arXiv preprint arXiv:2403.04811}, 2024.

\bibitem[Roberts et~al.(2023)Roberts, Thakur, Herlihy, White, and Dooley]{roberts2023data}
Manley Roberts, Himanshu Thakur, Christine Herlihy, Colin White, and Samuel Dooley.
\newblock Data contamination through the lens of time.
\newblock \emph{arXiv preprint arXiv:2310.10628}, 2023.

\bibitem[Sainz et~al.(2023)Sainz, Campos, Garc{\'\i}a-Ferrero, Etxaniz, de~Lacalle, and Agirre]{sainz2023nlp}
Oscar Sainz, Jon~Ander Campos, Iker Garc{\'\i}a-Ferrero, Julen Etxaniz, Oier~Lopez de~Lacalle, and Eneko Agirre.
\newblock Nlp evaluation in trouble: On the need to measure llm data contamination for each benchmark.
\newblock \emph{arXiv preprint arXiv:2310.18018}, 2023.

\bibitem[Sakaguchi et~al.(2021)Sakaguchi, Bras, Bhagavatula, and Choi]{winogrande}
Keisuke Sakaguchi, Ronan~Le Bras, Chandra Bhagavatula, and Yejin Choi.
\newblock Winogrande: An adversarial winograd schema challenge at scale.
\newblock \emph{Communications of the ACM}, 64\penalty0 (9):\penalty0 99--106, 2021.

\bibitem[Samuel et~al.(2024)Samuel, Zhou, and Zou]{samuel2024towards}
Vinay Samuel, Yue Zhou, and Henry~Peng Zou.
\newblock Towards data contamination detection for modern large language models: Limitations, inconsistencies, and oracle challenges.
\newblock \emph{arXiv preprint arXiv:2409.09927}, 2024.

\bibitem[Sanh et~al.(2019)Sanh, Debut, Chaumond, and Wolf]{sanh2019distilbert}
Victor Sanh, Lysandre Debut, Julien Chaumond, and Thomas Wolf.
\newblock Distilbert, a distilled version of bert: smaller, faster, cheaper and lighter.
\newblock \emph{arXiv preprint arXiv:1910.01108}, 2019.

\bibitem[Sap et~al.(2019)Sap, Rashkin, Chen, LeBras, and Choi]{sap2019socialiqa}
Maarten Sap, Hannah Rashkin, Derek Chen, Ronan LeBras, and Yejin Choi.
\newblock Socialiqa: Commonsense reasoning about social interactions.
\newblock \emph{arXiv preprint arXiv:1904.09728}, 2019.

\bibitem[Schulman et~al.(2017)Schulman, Wolski, Dhariwal, Radford, and Klimov]{schulman2017proximal}
John Schulman, Filip Wolski, Prafulla Dhariwal, Alec Radford, and Oleg Klimov.
\newblock Proximal policy optimization algorithms.
\newblock \emph{arXiv preprint arXiv:1707.06347}, 2017.

\bibitem[Shi et~al.(2024)Shi, Ajith, Xia, Huang, Liu, Blevins, Chen, and Zettlemoyer]{shi2024detecting}
Weijia Shi, Anirudh Ajith, Mengzhou Xia, Yangsibo Huang, Daogao Liu, Terra Blevins, Danqi Chen, and Luke Zettlemoyer.
\newblock Detecting pretraining data from large language models.
\newblock In \emph{The Twelfth International Conference on Learning Representations}, 2024.
\newblock URL \url{https://openreview.net/forum?id=zWqr3MQuNs}.

\bibitem[Singh et~al.(2024)Singh, Kocyigit, Poulton, Esiobu, Lomeli, Szilvasy, and Hupkes]{singh2024evaluation}
Aaditya~K Singh, Muhammed~Yusuf Kocyigit, Andrew Poulton, David Esiobu, Maria Lomeli, Gergely Szilvasy, and Dieuwke Hupkes.
\newblock Evaluation data contamination in llms: how do we measure it and (when) does it matter?
\newblock \emph{arXiv preprint arXiv:2411.03923}, 2024.

\bibitem[Srivastava et~al.(2022)Srivastava, Rastogi, Rao, Shoeb, Abid, Fisch, Brown, Santoro, Gupta, Garriga-Alonso, et~al.]{bigbench}
Aarohi Srivastava, Abhinav Rastogi, Abhishek Rao, Abu Awal~Md Shoeb, Abubakar Abid, Adam Fisch, Adam~R Brown, Adam Santoro, Aditya Gupta, Adri{\`a} Garriga-Alonso, et~al.
\newblock Beyond the imitation game: Quantifying and extrapolating the capabilities of language models.
\newblock \emph{arXiv preprint arXiv:2206.04615}, 2022.

\bibitem[Sun et~al.(2020)Sun, Wang, Chen, Ni, Agrawal, Cui, Venkataramani, El~Maghraoui, Srinivasan, and Gopalakrishnan]{sun2020ultra}
Xiao Sun, Naigang Wang, Chia-Yu Chen, Jiamin Ni, Ankur Agrawal, Xiaodong Cui, Swagath Venkataramani, Kaoutar El~Maghraoui, Vijayalakshmi~Viji Srinivasan, and Kailash Gopalakrishnan.
\newblock Ultra-low precision 4-bit training of deep neural networks.
\newblock \emph{Advances in Neural Information Processing Systems}, 33:\penalty0 1796--1807, 2020.

\bibitem[Team et~al.(2023)Team, Anil, Borgeaud, Wu, Alayrac, Yu, Soricut, Schalkwyk, Dai, Hauth, et~al.]{gemini}
Gemini Team, Rohan Anil, Sebastian Borgeaud, Yonghui Wu, Jean-Baptiste Alayrac, Jiahui Yu, Radu Soricut, Johan Schalkwyk, Andrew~M Dai, Anja Hauth, et~al.
\newblock Gemini: a family of highly capable multimodal models.
\newblock \emph{arXiv preprint arXiv:2312.11805}, 2023.

\bibitem[Touvron et~al.(2023{\natexlab{a}})Touvron, Lavril, Izacard, Martinet, Lachaux, Lacroix, Rozi{\`e}re, Goyal, Hambro, Azhar, et~al.]{llama1}
Hugo Touvron, Thibaut Lavril, Gautier Izacard, Xavier Martinet, Marie-Anne Lachaux, Timoth{\'e}e Lacroix, Baptiste Rozi{\`e}re, Naman Goyal, Eric Hambro, Faisal Azhar, et~al.
\newblock Llama: Open and efficient foundation language models.
\newblock \emph{arXiv preprint arXiv:2302.13971}, 2023{\natexlab{a}}.

\bibitem[Touvron et~al.(2023{\natexlab{b}})Touvron, Martin, Stone, Albert, Almahairi, Babaei, Bashlykov, Batra, Bhargava, Bhosale, et~al.]{touvron2023llama}
Hugo Touvron, Louis Martin, Kevin Stone, Peter Albert, Amjad Almahairi, Yasmine Babaei, Nikolay Bashlykov, Soumya Batra, Prajjwal Bhargava, Shruti Bhosale, et~al.
\newblock Llama 2: Open foundation and fine-tuned chat models.
\newblock \emph{arXiv preprint arXiv:2307.09288}, 2023{\natexlab{b}}.

\bibitem[Tu et~al.(2024)Tu, Zhu, Bai, Yao, Hou, and Li]{tu2024dice}
Shangqing Tu, Kejian Zhu, Yushi Bai, Zijun Yao, Lei Hou, and Juanzi Li.
\newblock Dice: Detecting in-distribution contamination in llm's fine-tuning phase for math reasoning.
\newblock \emph{arXiv preprint arXiv:2406.04197}, 2024.

\bibitem[Wang et~al.(2024)Wang, Bao, Wu, Taylor, Xiao, Zheng, Jiang, Gao, and Zhang]{wang2024unlock-memorizing}
Zhepeng Wang, Runxue Bao, Yawen Wu, Jackson Taylor, Cao Xiao, Feng Zheng, Weiwen Jiang, Shangqian Gao, and Yanfu Zhang.
\newblock Unlocking memorization in large language models with dynamic soft prompting.
\newblock In Yaser Al{-}Onaizan, Mohit Bansal, and Yun{-}Nung Chen (eds.), \emph{Proceedings of the 2024 Conference on Empirical Methods in Natural Language Processing, {EMNLP} 2024, Miami, FL, USA, November 12-16, 2024}, pp.\  9782--9796. Association for Computational Linguistics, 2024.
\newblock URL \url{https://aclanthology.org/2024.emnlp-main.546}.

\bibitem[Wei et~al.(2022)Wei, Wang, Schuurmans, Bosma, Xia, Chi, Le, Zhou, et~al.]{wei2022chain}
Jason Wei, Xuezhi Wang, Dale Schuurmans, Maarten Bosma, Fei Xia, Ed~Chi, Quoc~V Le, Denny Zhou, et~al.
\newblock Chain-of-thought prompting elicits reasoning in large language models.
\newblock \emph{Advances in neural information processing systems}, 35:\penalty0 24824--24837, 2022.

\bibitem[Weller et~al.(2023)Weller, Marone, Weir, Lawrie, Khashabi, and Van~Durme]{weller2023according}
Orion Weller, Marc Marone, Nathaniel Weir, Dawn Lawrie, Daniel Khashabi, and Benjamin Van~Durme.
\newblock " according to..." prompting language models improves quoting from pre-training data.
\newblock \emph{arXiv preprint arXiv:2305.13252}, 2023.

\bibitem[White et~al.(2024)White, Dooley, Roberts, Pal, Feuer, Jain, Shwartz-Ziv, Jain, Saifullah, Naidu, et~al.]{white2024livebench}
Colin White, Samuel Dooley, Manley Roberts, Arka Pal, Ben Feuer, Siddhartha Jain, Ravid Shwartz-Ziv, Neel Jain, Khalid Saifullah, Siddartha Naidu, et~al.
\newblock Livebench: A challenging, contamination-free llm benchmark.
\newblock \emph{arXiv preprint arXiv:2406.19314}, 2024.

\bibitem[Witteveen \& Andrews(2019)Witteveen and Andrews]{witteveen2019paraphrasing}
Sam Witteveen and Martin Andrews.
\newblock Paraphrasing with large language models.
\newblock \emph{arXiv preprint arXiv:1911.09661}, 2019.

\bibitem[Wu et~al.(2023)Wu, Bansal, Zhang, Wu, Zhang, Zhu, Li, Jiang, Zhang, and Wang]{wu2023autogen}
Qingyun Wu, Gagan Bansal, Jieyu Zhang, Yiran Wu, Shaokun Zhang, Erkang Zhu, Beibin Li, Li~Jiang, Xiaoyun Zhang, and Chi Wang.
\newblock Autogen: Enabling next-gen llm applications via multi-agent conversation framework.
\newblock \emph{arXiv preprint arXiv:2308.08155}, 2023.

\bibitem[Xie et~al.(2023)Xie, Han, Zhang, Lai, Peng, Lopez-Lira, and Huang]{xie2023pixiu}
Qianqian Xie, Weiguang Han, Xiao Zhang, Yanzhao Lai, Min Peng, Alejandro Lopez-Lira, and Jimin Huang.
\newblock Pixiu: A large language model, instruction data and evaluation benchmark for finance.
\newblock \emph{arXiv preprint arXiv:2306.05443}, 2023.

\bibitem[Xie et~al.(2024)Xie, Wang, Huang, Zhang, Ge, Pei, Gong, and Dhingra]{xie2024recall}
Roy Xie, Junlin Wang, Ruomin Huang, Minxing Zhang, Rong Ge, Jian Pei, Neil~Zhenqiang Gong, and Bhuwan Dhingra.
\newblock Recall: Membership inference via relative conditional log-likelihoods.
\newblock \emph{arXiv preprint arXiv:2406.15968}, 2024.

\bibitem[Xu et~al.(2024{\natexlab{a}})Xu, Guan, Greene, Kechadi, et~al.]{xu2024benchmark}
Cheng Xu, Shuhao Guan, Derek Greene, M~Kechadi, et~al.
\newblock Benchmark data contamination of large language models: A survey.
\newblock \emph{arXiv preprint arXiv:2406.04244}, 2024{\natexlab{a}}.

\bibitem[Xu et~al.(2024{\natexlab{b}})Xu, Wang, Fan, and Liu]{xu2024benchmarking}
Ruijie Xu, Zengzhi Wang, Run-Ze Fan, and Pengfei Liu.
\newblock Benchmarking benchmark leakage in large language models.
\newblock \emph{arXiv preprint arXiv:2404.18824}, 2024{\natexlab{b}}.

\bibitem[Yang et~al.(2023{\natexlab{a}})Yang, Xiao, Wang, Zhang, Bian, Yin, Lv, Pan, Wang, Yan, et~al.]{yang2023baichuan}
Aiyuan Yang, Bin Xiao, Bingning Wang, Borong Zhang, Ce~Bian, Chao Yin, Chenxu Lv, Da~Pan, Dian Wang, Dong Yan, et~al.
\newblock Baichuan 2: Open large-scale language models.
\newblock \emph{arXiv preprint arXiv:2309.10305}, 2023{\natexlab{a}}.

\bibitem[Yang et~al.(2023{\natexlab{b}})Yang, Yue, and He]{yang2023auto}
Hui Yang, Sifu Yue, and Yunzhong He.
\newblock Auto-gpt for online decision making: Benchmarks and additional opinions.
\newblock \emph{arXiv preprint arXiv:2306.02224}, 2023{\natexlab{b}}.

\bibitem[Yang et~al.(2023{\natexlab{c}})Yang, Chiang, Zheng, Gonzalez, and Stoica]{yang2023rethinking}
Shuo Yang, Wei-Lin Chiang, Lianmin Zheng, Joseph~E Gonzalez, and Ion Stoica.
\newblock Rethinking benchmark and contamination for language models with rephrased samples.
\newblock \emph{arXiv preprint arXiv:2311.04850}, 2023{\natexlab{c}}.

\bibitem[Young et~al.(2024)Young, Chen, Li, Huang, Zhang, Zhang, Li, Zhu, Chen, Chang, et~al.]{young2024yi}
Alex Young, Bei Chen, Chao Li, Chengen Huang, Ge~Zhang, Guanwei Zhang, Heng Li, Jiangcheng Zhu, Jianqun Chen, Jing Chang, et~al.
\newblock Yi: Open foundation models by 01. ai.
\newblock \emph{arXiv preprint arXiv:2403.04652}, 2024.

\bibitem[Yu et~al.(2018)Yu, Zhang, Yang, Yasunaga, Wang, Li, Ma, Li, Yao, Roman, et~al.]{yu2018spider}
Tao Yu, Rui Zhang, Kai Yang, Michihiro Yasunaga, Dongxu Wang, Zifan Li, James Ma, Irene Li, Qingning Yao, Shanelle Roman, et~al.
\newblock Spider: A large-scale human-labeled dataset for complex and cross-domain semantic parsing and text-to-sql task.
\newblock \emph{arXiv preprint arXiv:1809.08887}, 2018.

\bibitem[Yu et~al.(2024)Yu, Gao, Yao, Wang, Ye, Wang, Xie, Zhang, and Zhang]{yu2024kieval}
Zhuohao Yu, Chang Gao, Wenjin Yao, Yidong Wang, Wei Ye, Jindong Wang, Xing Xie, Yue Zhang, and Shikun Zhang.
\newblock Kieval: A knowledge-grounded interactive evaluation framework for large language models, 2024.

\bibitem[Zellers et~al.(2019)Zellers, Holtzman, Bisk, Farhadi, and Choi]{hellaswag}
Rowan Zellers, Ari Holtzman, Yonatan Bisk, Ali Farhadi, and Yejin Choi.
\newblock Hellaswag: Can a machine really finish your sentence?
\newblock \emph{arXiv preprint arXiv:1905.07830}, 2019.

\bibitem[Zhang \& Wu(2024)Zhang and Wu]{zhang2024adaptive}
Anqi Zhang and Chaofeng Wu.
\newblock Adaptive pre-training data detection for large language models via surprising tokens.
\newblock \emph{arXiv preprint arXiv:2407.21248}, 2024.

\bibitem[Zhang et~al.(2024{\natexlab{a}})Zhang, Sun, Yeats, Ouyang, Kuo, Zhang, Yang, and Li]{zhang2024min}
Jingyang Zhang, Jingwei Sun, Eric Yeats, Yang Ouyang, Martin Kuo, Jianyi Zhang, Hao~Frank Yang, and Hai Li.
\newblock Min-k\%++: Improved baseline for detecting pre-training data from large language models.
\newblock \emph{arXiv preprint arXiv:2404.02936}, 2024{\natexlab{a}}.

\bibitem[Zhang et~al.(2023)Zhang, Cai, Liu, Yang, Dai, Liao, Qin, Li, Liu, Liu, et~al.]{zhang2023fineval}
Liwen Zhang, Weige Cai, Zhaowei Liu, Zhi Yang, Wei Dai, Yujie Liao, Qianru Qin, Yifei Li, Xingyu Liu, Zhiqiang Liu, et~al.
\newblock Fineval: A chinese financial domain knowledge evaluation benchmark for large language models.
\newblock \emph{arXiv preprint arXiv:2308.09975}, 2023.

\bibitem[Zhang et~al.(2024{\natexlab{b}})Zhang, Zhang, Guo, de~Rijke, Fan, and Cheng]{zhang2024pretraining}
Weichao Zhang, Ruqing Zhang, Jiafeng Guo, Maarten de~Rijke, Yixing Fan, and Xueqi Cheng.
\newblock Pretraining data detection for large language models: A divergence-based calibration method.
\newblock \emph{arXiv preprint arXiv:2409.14781}, 2024{\natexlab{b}}.

\bibitem[Zhang et~al.(2015)Zhang, Zhao, and LeCun]{agnews}
Xiang Zhang, Junbo Zhao, and Yann LeCun.
\newblock Character-level convolutional networks for text classification.
\newblock \emph{Advances in neural information processing systems}, 28, 2015.

\bibitem[Zhao et~al.(2024)Zhao, Li, and Yang]{zhao2024cap}
Yi~Zhao, Jing Li, and Linyi Yang.
\newblock Cap: Data contamination detection via consistency amplification.
\newblock \emph{arXiv preprint arXiv:2410.15005}, 2024.

\bibitem[Zheng et~al.(2023)Zheng, Chiang, Sheng, Zhuang, Wu, Zhuang, Lin, Li, Li, Xing, et~al.]{zheng2023judging}
Lianmin Zheng, Wei-Lin Chiang, Ying Sheng, Siyuan Zhuang, Zhanghao Wu, Yonghao Zhuang, Zi~Lin, Zhuohan Li, Dacheng Li, Eric Xing, et~al.
\newblock Judging llm-as-a-judge with mt-bench and chatbot arena.
\newblock \emph{Advances in Neural Information Processing Systems}, 36:\penalty0 46595--46623, 2023.

\bibitem[Zhong et~al.(2023)Zhong, Cui, Guo, Liang, Lu, Wang, Saied, Chen, and Duan]{agieval}
Wanjun Zhong, Ruixiang Cui, Yiduo Guo, Yaobo Liang, Shuai Lu, Yanlin Wang, Amin Saied, Weizhu Chen, and Nan Duan.
\newblock Agieval: A human-centric benchmark for evaluating foundation models.
\newblock \emph{arXiv preprint arXiv:2304.06364}, 2023.

\bibitem[Zhou et~al.(2023)Zhou, Zhu, Chen, Chen, Zhao, Chen, Lin, Wen, and Han]{zhou2023don}
Kun Zhou, Yutao Zhu, Zhipeng Chen, Wentong Chen, Wayne~Xin Zhao, Xu~Chen, Yankai Lin, Ji-Rong Wen, and Jiawei Han.
\newblock Don't make your llm an evaluation benchmark cheater.
\newblock \emph{arXiv preprint arXiv:2311.01964}, 2023.

\bibitem[Zhu et~al.(2024)Zhu, Cheng, Peng, Li, Liu, Peng, Qiu, and Huang]{zhu2024inference}
Qin Zhu, Qingyuan Cheng, Runyu Peng, Xiaonan Li, Tengxiao Liu, Ru~Peng, Xipeng Qiu, and Xuanjing Huang.
\newblock Inference-time decontamination: Reusing leaked benchmarks for large language model evaluation.
\newblock \emph{arXiv preprint arXiv:2406.13990}, 2024.

\bibitem[Zhu et~al.(2023)Zhu, Hao, He, Song, Zhang, Hu, Wei, Wang, and Lu]{zhu2023clean}
Wenhong Zhu, Hongkun Hao, Zhiwei He, Yunze Song, Yumeng Zhang, Hanxu Hu, Yiran Wei, Rui Wang, and Hongyuan Lu.
\newblock Clean-eval: Clean evaluation on contaminated large language models.
\newblock \emph{arXiv preprint arXiv:2311.09154}, 2023.

\end{thebibliography}
